%% file: neurips_2026.tex
\documentclass{article}

\PassOptionsToPackage{numbers,sort&compress}{natbib}
\usepackage[preprint]{neurips_2026}

\makeatletter
\renewcommand{\@notice}{}
\makeatother

\usepackage{amsmath}
\usepackage[utf8]{inputenc}
\usepackage[T1]{fontenc}
\usepackage{hyperref}
\usepackage[capitalize,nameinlink]{cleveref}
\usepackage{url}
\usepackage{graphicx}
\usepackage{booktabs}
\usepackage{multirow}
\usepackage{makecell}
\usepackage{siunitx}
\usepackage{xcolor}

\usepackage{amssymb}
\usepackage{amsfonts}
\usepackage{nicefrac}
\usepackage{microtype}
\usepackage{pifont}
\usepackage[ruled,vlined,linesnumbered]{algorithm2e}
\usepackage[most]{tcolorbox}
\usepackage{enumitem}
\usepackage{eccvabbrv}

\newcommand{\cmark}{\textcolor{green!60!black}{\ding{51}}}
\newcommand{\xmark}{\textcolor{red!70!black}{\ding{55}}}
\newcommand{\keywords}[1]{%
  \par\smallskip\noindent\textbf{Keywords:}%
  \begingroup\renewcommand{\and}{;\ } #1\endgroup%
}

\title{Learning Explicit Physical Parameter Control and Benchmarking for Video Generation}

\author{%
  \normalfont
  Yanxun Li\textsuperscript{1}\thanks{Work done during the internship at AMAP.},
  Hao Wen\textsuperscript{2},
  Bingze Song\textsuperscript{2},
  Jiashu Zhu\textsuperscript{2},
  Aiming Hao\textsuperscript{2},
  \\
  Chubin Chen\textsuperscript{2},
  Jintao Chen\textsuperscript{2},
  Jiahong Wu\textsuperscript{2}\thanks{Project Leader.},
  Xiangxiang Chu\textsuperscript{2},
  Miao Wang\textsuperscript{1}\thanks{Corresponding author.}
  \\
  \textsuperscript{1}State Key Laboratory of Virtual Reality Technology and Systems, SCSE, Beihang University
  \\
  \textsuperscript{2}AMAP, Alibaba Group
  \\
  \texttt{yanxunli@buaa.edu.cn}
  \quad
  \texttt{hongxi.wjh@alibaba-inc.com}
}

\begin{document}

\maketitle

\input{sec/sec0}

\input{sec/sec1_2}

\input{sec/sec3}

\input{sec/sec4_method}

\input{sec/sec5_experiment}

\input{sec/sec6_conclusion}

\bibliographystyle{plainnat}
\bibliography{main}

\appendix
\input{sec/X_suppl}

\end{document}

%% file: sec/sec0.tex
\begin{abstract}
    % Recent advances in video generation have enhanced visual realism, opening new possibilities for constructing ``world simulators'' that capture real-world dynamics. However, a key challenge remains—can these models truly learn and obey physical laws? 
    Recent advances in image-to-video generation have improved visual realism, making physically grounded and controllable dynamics an important step toward future world simulation.
    Current models often generate plausible motion, but it is not reliably governed by explicit physical causes, and instance-level constraints can leak or become entangled in multi-object interactions.
    We attribute this gap to two missing pieces: large-scale, fine-grained physical parameterization, and model designs that correctly bind physical attributes to instances and emphasize dynamics over appearance.
    % To bridge the gap between generative modeling and physical signals, we introduce the \textbf{PhyParam-Dataset}, an interaction-centric collection of 130K physically simulated videos with fine-grained annotations of forces, object properties, and scene parameters across six representative object–object and object–environment interactions. 
    To bridge this gap, we introduce \textbf{PhyParam-Dataset}, an interaction-centric collection of 130K physically simulated videos with dense physical parameterization, including force vectors, object material properties, and environmental constants across five representative rigid-body motion types.
    Built on this data, we present \textbf{PhyParam}, a physics-guided image-to-video diffusion model that conditions on object-level forces, masses, friction, restitution, and scene-level gravity via a lightweight physical-attention routing mechanism, and further improves motion learning with semantic-structural feature-space supervision.
    We also establish \textbf{PhyParam-Bench}, a benchmark for physical-law consistency in image-to-video generation, with a multi-level protocol evaluating temporal dynamics, spatial stability, and semantic--physical alignment.
    % song note: 这里需要注意一下描述：force vectors, object material properties, and environmental constants
    % song note: 这里也需要注意一下描述，把PhyForce改成PhyParam
    Experiments show that PhyParam improves physical consistency while maintaining high visual fidelity, advancing explicit rigid-body physical-parameter control for image-to-video generation.
    We will publicly release the dataset, benchmark, and code to support future research.
  \keywords{Video generation \and Physical parameter control \and Physics consistency}
\end{abstract}

%% file: sec/sec1_2.tex
\section{Introduction}
\label{sec:intro}

The impressive progress of video generation raises a long-term question: \textit{How can generative models move from visually plausible videos toward physically grounded world simulation?}
Recent image-to-video (I2V) models have made rapid progress in temporal coherence and perceptual realism~\cite{wan2.1,kong2024hunyuanvideo,zheng2024open,peng2025open}.
They can also produce intuitive motion in common scenarios, suggesting that parts of physical regularities may be implicitly captured from data~\cite{shi2024motion}.

However, this simulation capability is often tied to frequent parameter regimes in the training distribution and does not guarantee robust physical generalization~\cite{liu2022compositional, huang2026taming}.
When encountering edge-case physical settings---such as extreme velocities, unconventional material properties, or complex multi-object interactions---these models often yield physically implausible results.
Furthermore, current architectures lack fine-grained controllability, making it difficult to precisely steer physical attributes beyond learned statistical averages.
Consequently, a substantial gap persists between visually plausible video generation and explicit, controllable modeling of rigid-body physical parameters~\cite{bansal2024videophy, meng2024towards, wang2025wisa}.

\begin{figure}
    \centering
    \vspace{-10pt}
    \includegraphics[width=0.95\linewidth]{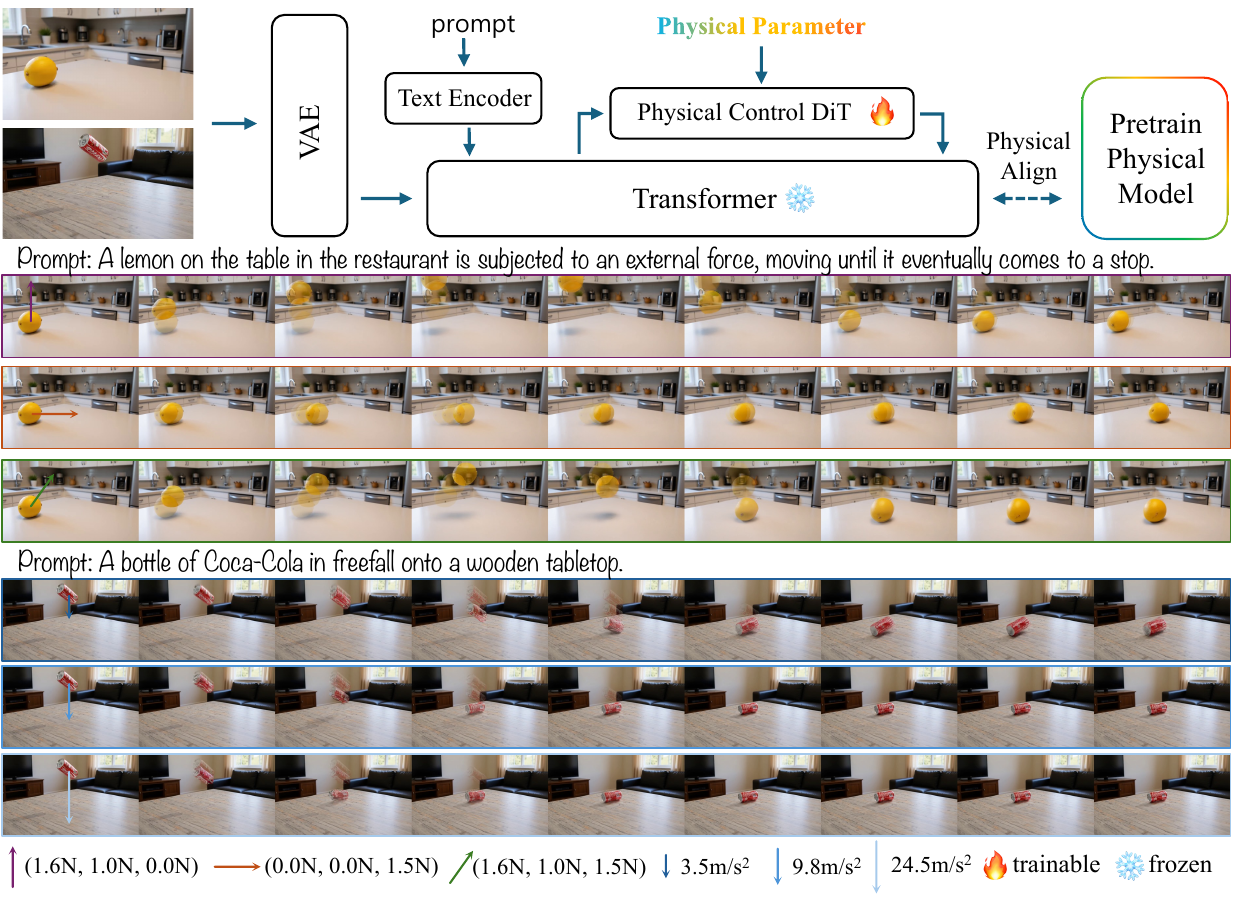}
    \vspace{-10pt}
    \caption{PhyParam, trained on the PhyParam-Dataset, enables precise and fine-grained control over physical parameters. The figure illustrates local force interventions applied from multiple directions, and varying global gravity configurations.}
    \label{fig:teasor}
\end{figure}

% Unfortunately, while current image-to-video (I2V) models~\cite{wan2.1,kong2024hunyuanvideo,shi2024motion,zheng2024open,peng2025open} excel at capturing appearance and motion continuity, they often violate basic physical principles—objects may accelerate without force, penetrate each other, or float unnaturally against gravity. Although recent efforts introduce physics-inspired constraints, most remain confined to simple, single-object scenarios and struggle to model the intricate interactive dynamics that characterize real-world physical systems~\cite{bansal2024videophy, meng2024towards, wang2025wisa}. Consequently, a substantial gap persists between visual realism and physically grounded world simulation.

% Motivated by the need for more controllable and physically accurate world simulators, we present \textbf{PhyParam-Dataset}, the first interaction-centric dataset comprising 130K physically simulated videos with fine-grained parameter annotations. Built on the Blender~\footnote{https://www.blender.org/} physics engine, the dataset systematically simulates diverse combinations of physical parameters—such as gravity, friction, and restitution—across six representative object–object and object–environment interactions, including free fall, elastic and inelastic collisions, inclined-plane motion, and external-force-driven dynamics. Each video is annotated with explicit physical parameters and state trajectories, enabling supervised learning and quantitative evaluation of physical consistency in generative video models.

Motivated by the need for more controllable and physically grounded rigid-body dynamics, we present \textbf{PhyParam-Dataset}, an interaction-centric dataset designed to push generative models toward the long-tail of physical distributions through fine-grained parameter control.
Built with Blender,\footnote{https://www.blender.org/} our dataset comprises 130K high-fidelity simulated videos that systematically explore diverse combinations of fundamental physical parameters, including gravity, friction, and restitution.
The dataset spans five representative rigid-body motion types: free fall, elastic collision, inelastic collision, inclined-plane motion, and external-force-driven dynamics.
% 这里需要注意一下这几个专有名词，force包括 gravity, friction, and restitution；interaction 包括 free fall, elastic and inelastic collisions, inclined-plane motion, and external-force-driven dynamics.

Leveraging the fine-grained physical signals provided by our dataset, we propose \textbf{PhyParam}, as shown in Fig.~\ref{fig:teasor}, a physics-guided generative model that combines explicit physical priors with a novel conditioning and supervision design.
PhyParam encodes global and local physical parameters---including gravity, object-specific forces, and material properties---into a unified representation via a lightweight physical-attention module, enabling precise multi-condition control without significant computational overhead.
Furthermore, a semantic-structural feature-space supervision scheme aligns generated object identity, boundaries, and temporal displacement with reference physical simulations, improving physical-law compliance without assuming that generic visual features directly encode physical variables.

Beyond generation frameworks, existing video evaluation benchmarks predominantly emphasize perceptual and semantic fidelity~\cite{huang2024vbench,liu2024physgen,liu2024evalcrafter,zhu2026artifact}.
Recent efforts begin to evaluate physical commonsense and dynamics, but they are still largely confined to common real-world regimes, making it difficult to diagnose physical consistency under long-tail settings~\cite{bansal2024videophy,bansal2025videophy,meng2024towards,zhang2025morpheus,motamed2025generative}.
To bridge this gap, we introduce \textbf{PhyParam-Bench}, a benchmark tailored for evaluating \emph{Physical Law Consistency} in rigid-body motion.
This bench serves as a practical diagnostic tool by employing a three-level evaluation protocol centered on temporal dynamics, spatial stability, and semantic--physical alignment.
Extensive evaluation on the bench and user studies demonstrates that our method achieves strong physical consistency while preserving perceptual quality.

Together, we present a closed-loop pipeline for explicit instance-level physical control in I2V generation: a parameterized dataset, a physics-guided generator, and a dedicated benchmark:
\begin{itemize}
    \item \textbf{PhyParam-Dataset}: 130K physics-simulated interaction videos with parameter and trajectory annotations for controllable dynamics learning.
    \item \textbf{PhyParam}: a physics-guided diffusion model with parameter conditioning, improving controllability and physical consistency.
    \item \textbf{PhyParam-Bench}: 380-test benchmark with a hierarchical protocol to evaluate temporal dynamics, spatial stability, and semantic--physical alignment under long-tail regimes.
\end{itemize}

\section{Related Works}

\subsection{Physical-aware Video Generation}
The long-term goal of using video generation models as components of world simulators has gained momentum with recent advances in large-scale generative models~\cite{sora, sora2, hong2022cogvideo,kong2024hunyuanvideo,wan2.1,zhao2026attention}. This vision is closely tied to the long-standing pursuit of embedding physical understanding into generative models. Early attempts modeled motion as 2D image-space vibrations~\cite{davis2015visual, davis2015visual2, wang2025wisa}, which handled oscillatory behaviors, but failed on more general dynamics.

A major research thread introduces physics simulators to enforce physical plausibility. Some methods reconstruct 3D scene representations~\cite{kerbl20233d,mildenhall2021nerf} and apply simulation before rendering~\cite{wang2024motionctrl,shi2024motion,zhang2025tora}. Although physically faithful, these approaches are scene-specific and require high-quality 3D reconstruction. Hybrid approaches such as PhysGen~\cite{liu2024physgen}, PhysMotion~\cite{tan2024physmotion}, and PhysAnimator~\cite{xie2025physanimator} instead use a simulator to generate object trajectories and then rely on a video generator for rendering. However, they are heavily dependent on an external simulator, limiting dynamic diversity and demanding extensive engineering and tuning.

Recent work moves toward embedding physical priors directly into generative models. Some leverage auxiliary signals—e.g., LLM-based causal inference~\cite{li2025c} or control masks that were unavailable at test time~\cite{akkerman2025interdyn}. Other concurrent efforts fine-tune on simulated data~\cite{li2025pisa, romero2025learning} but focus on simple phenomena such as gravity or free fall. Force-based conditioning has also been explored~\cite{gillman2025force, li2025wonderplay}, although often without addressing complex interaction of objects.

Distinct from simulator-dependent approaches, our method integrates physics priors directly into a diffusion model~\cite{peebles2023scalable}. This removes the need for a simulator at inference time, improves numerical stability, and enables controllable, physically grounded video generation with more complex dynamics.

\subsection{Physical Video Generation Benchmark}

\textbf{Benchmarks for Video Generation.} Evaluation of text-to-video models has progressed with comprehensive benchmarks that assess diverse aspects of generation quality. Frameworks such as VBench~\cite{huang2024vbench,zheng2025vbench,huang2024vbench++},
EvalCrafter~\cite{liu2024evalcrafter}, T2V-CompBench~\cite{sun2025t2v},
and other benchmarks~\cite{ling2025vmbench, liu2026towards,feng2025narrlv} provide metrics for visual
fidelity, temporal coherence, and compositional alignment. However, these benchmarks mainly focus on perceptual and semantic quality, while largely overlooking the physical plausibility of generated dynamics. To address this gap, a recent line of work has focused on developing benchmarks specifically for physical reasoning. These include VideoPhy~\cite{bansal2024videophy, bansal2025videophy}, which systematically evaluates object interactions like collisions; PhyGenBench~\cite{meng2024towards}, which tests physical commonsense; and Physics-IQ~\cite{motamed2025generative}, which assesses a model's ability to predict future frames based on physical principles. Other notable efforts include DEVIL~\cite{liao2024evaluation} for evaluating dynamic realism and Morpheus~\cite{zhang2025morpheus} for physical plausibility. While these benchmarks provide valuable insights, they often focus on specific phenomena or do not cover a sufficiently broad range of physical laws, highlighting the need for more comprehensive evaluation.

Evaluation Metrics. Traditional metrics like Frechet Video Distance (FVD)~\cite{unterthiner2018towards} are insufficient for this task, as they primarily measure visual quality and fail to capture the correctness of physical motion, especially in reference-free scenarios. Consequently, recent benchmarks have introduced more targeted evaluation strategies. These range from automated metrics like Intersection over Union (IoU) for geometric consistency~\cite{motamed2025generative} to leveraging large vision-language models (VLMs)~\cite{Qwen-VL,Qwen2-VL,Qwen2.5-VL} for semantic and compositional assessment~\cite{sun2025t2v}. Furthermore, to capture nuanced aspects of physical realism, many benchmarks rely on human evaluation~\cite{he2024videoscore, bansal2025videophy} or introduce efficient, scalable answering schemes to objectively assess physical commonsense in generated videos.

%% file: sec/sec3.tex
\section{Dataset and Benchmark}
\label{sec:dataset}

\begin{figure*}[t!]
    \centering
    \includegraphics[width=0.95\linewidth]{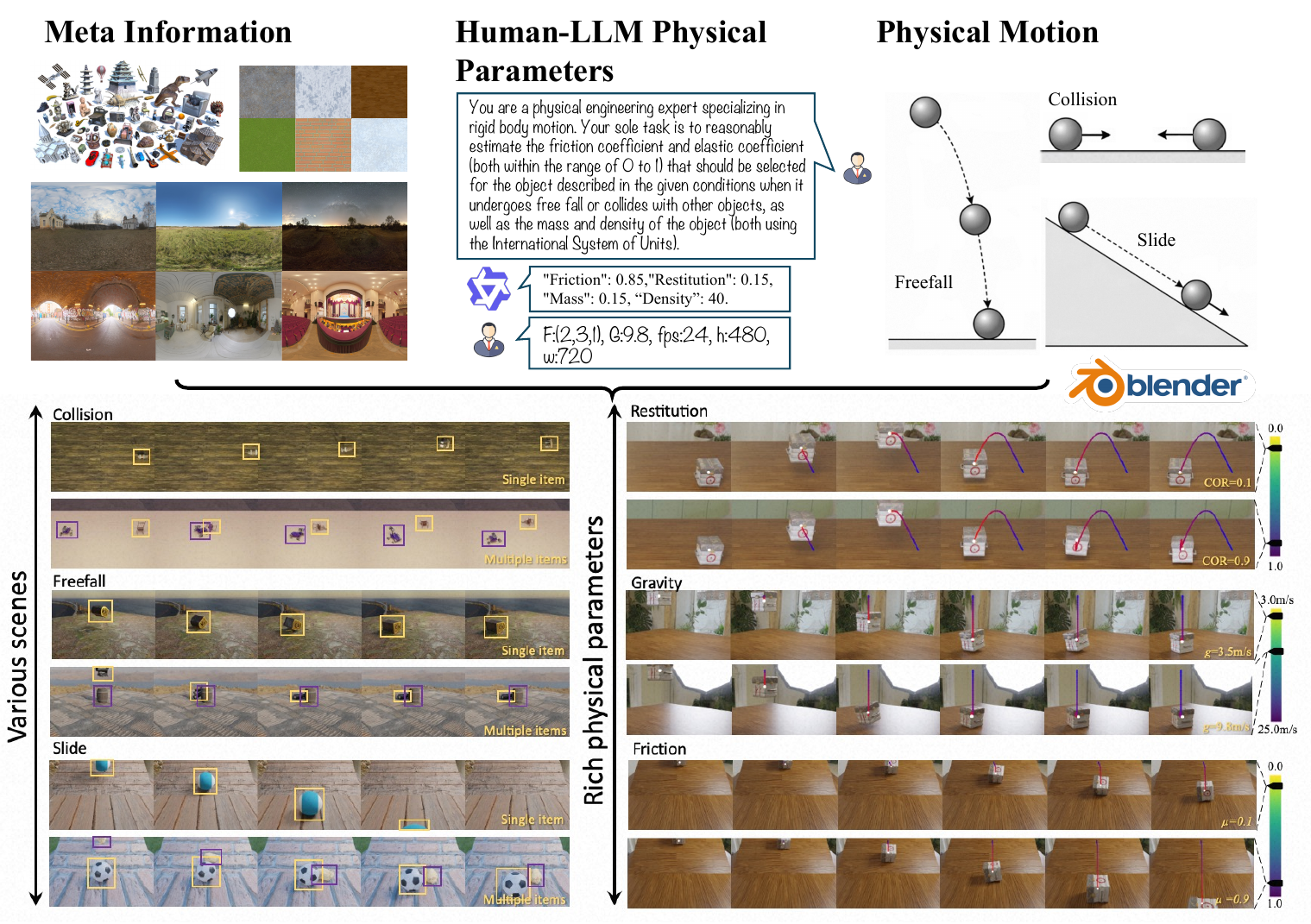}
    \vspace{-10pt}
    \caption{Overview of the PhyParam-Dataset generation pipeline. Our pipeline constructs physics-consistent video--text pairs by combining multi-source assets, LLM-predicted physical parameters, and physics-based simulation. }
    \label{fig:dataset}
\end{figure*}
% \vspace{-20pt}
% Specifically, we assemble 3D assets from Objaverse~\cite{deitke2023objaverse} with materials from CC-Textures~\cite{denninger2023blenderproc2} and HDRI environments from PolyHaven~\cite{haven2025public}, then apply predefined motion templates (free fall, collision, and sliding). We use Qwen~\cite{Qwen2.5-VL} to infer object-specific physical parameters (mass, friction, and restitution) and to generate descriptive captions. Finally, Blender simulates and renders the scenes into videos, along with per-instance segmentation masks and physical-parameter annotations, yielding high-quality training examples for physics-aware video generation.
% Despite growing interest in physical consistency, existing video generation models largely remain confined to isolated single-object motion, failing to capture the interdependent dynamics that govern real-world scenes. To move beyond this limitation, we introduce PhyParam--a unified framework featuring an interaction-centric dataset with explicit physical annotations and a benchmark designed to evaluate complex, multi-object physical dynamics.
Despite growing interest in physical consistency, existing video generation models still focus on isolated single-object motion and fail to capture the interdependent dynamics of real-world scenes. 
One major reason is the lack of large-scale datasets with explicit physical annotations. To move beyond this limitation, we introduce a unified framework combining an interaction-centric dataset and a benchmark for complex, multi-object physical dynamics.

\subsection{PhyParam Dataset}

\noindent\textbf{Construction Pipeline.} As shown in ~\cref{fig:dataset}, we construct PhyParam-Dataset through a scalable and reproducible pipeline based on the Blender physics engine. 
We sample diverse 3D objects from Objaverse~\cite{deitke2023objaverse}, build planar grounds with CC-Textures~\cite{denninger2023blenderproc2}, and use HDRI maps from PolyHaven~\cite{haven2025public} to provide realistic lighting and reflections. 
The pipeline follows scene design and rendering strategies from ReCamMaster~\cite{bai2025recammaster} and MirrorVerse~\cite{dhiman2025mirrorverse}.
Physical simulations are configured with continuous parameter variations, including gravity, object mass, external force, restitution, and friction coefficient, to ensure controllable diversity.  
To guarantee data quality, all videos undergo automated and manual validation for parameter accuracy and physical consistency.
Over 98\% satisfy force-balance and smooth-trajectory constraints, confirming PhyParam-Dataset as a scalable, verifiable, and physically grounded resource for physics-consistent video generation.
% The overall dataset construction process draws inspiration from the scene design and rendering strategies introduced in ReCamMaster~\cite{bai2025recammaster} and MirrorVerse~\cite{dhiman2025mirrorverse}. Each physical simulation is configured with continuous parameter variations ($e.g.$, gravity, restitution, and friction coefficient), ensuring controllable diversity across samples. We further leverage Qwen to expand pre-defined templates into paired physics-aware captions. 
% All videos are automatically and manually validated for parameter accuracy and physical consistency. Over 98\% satisfy the force-balance and smooth-trajectory constraints, confirming PhyParam-Dataset as a scalable, verifiable, and physically grounded resource for studying physically consistent video generation. More details can be seen in \textcolor{red}{Appendix}.

\noindent\textbf{Dataset Annotation.} Each sample is annotated with fine-grained physical parameters, including gravity, object mass, initial external force magnitude and direction, friction coefficient, and restitution coefficient.  
We also provide physics-aware textual descriptions, which are generated by expanding pre-defined templates using Qwen3-VL, 
first-frame segmentation masks, and frame-level physical state labels.  
These annotations form a complete alignment from low-level dynamic constraints to high-level semantic descriptions, 
establishing a quantitative and supervised foundation for training and evaluation, and enabling generative models to synthesize motion aligned with real-world physical laws.
% This design ensures the authenticity and continuity of physical parameters, offering a quantitative and supervised basis for both model training and evaluation. Consequently, it enables generative models to learn and synthesize motion sequences that comply with real-world physical laws.
\begin{table}[t!]
\centering
\small
\setlength{\tabcolsep}{3pt} % 列间距
\renewcommand{\arraystretch}{1} % 行距
\caption{Comparison of datasets containing physical parameters, including number of prompts \& videos, available attributes, and data type.} % 修改成你需要的标题
\label{tab:phy_dataset_stat}
\resizebox{0.60\linewidth}{!}{\begin{tabular}{lccccc}
\toprule
\multirow{2}*{\textbf{Dataset}} & \multirow{2}*{\textbf{Prompt\&Video}} &
\multicolumn{2}{c}{\textbf{Params}} & \multirow{2}*{\textbf{Type}} \\
\cmidrule(lr){3-4}
&  & \textbf{Object Attr.} & \textbf{Force} &  \\
\midrule
PhyBench         & 700        & \xmark & \xmark & generated \\
VideoPhy         & 688        & \xmark & \cmark & generated \\
PisaBench        & 361        & \xmark & \xmark & real \\
WISA-32K         & 32,000     & \cmark & \xmark & real \\
\textbf{PhyParam-Dataset}& \textbf{130,000} & \cmark & \cmark & \textbf{synthetic} \\
\bottomrule
\end{tabular}}
\end{table}

\noindent\textbf{Statistical Analysis.} As shown in \cref{tab:phy_dataset_stat}, compared with other datasets, the PhyParam Dataset is a large-scale resource for physics-consistent video generation, containing approximately $130$K text–video pairs. It consists of $64$K 3D objects, $944$ background environments, and $139$ ground materials, offering high diversity in both visual and physical composition.  
The dataset systematically covers three categories of physical scenarios—single-body motion, two-body collision, and multi-body dynamic interaction—including representative processes such as free fall, elastic and inelastic collisions, inclined-plane motion, and externally driven dynamics.
To reduce distribution bias, we perform balanced sampling across multiple physical dimensions (height, direction, interaction type), ensuring comprehensive coverage across diverse motion patterns. Additional analyses are included in the Appendix~\ref{sec:appendix_dataset}.

\subsection{PhyParam-Bench}
PhyParam-Bench contains 380 test videos across diverse physical motions, evenly distributed over different physical parameters and motion complexities. 
It is designed to evaluate \emph{Physical Law Consistency} (PLC), i.e., how well generative models comply with real-world physical constraints. 
The evaluation covers three main perspectives: \emph{Temporal dynamics}, \emph{Spatial structural stability}, and \emph{Semantic–physical alignment}. 
A summary of all metrics is provided in \cref{tab:metrics}, and we employ them to systematically analyze model performance across varying physical conditions.

\begin{table*}[ht]
\centering
\scriptsize
\caption{Overview of evaluation metrics in PhyParam-Bench}
\begin{tabular}{p{0.16\linewidth}|p{0.12\linewidth}|p{0.65\linewidth}}
\toprule
\textbf{Category} & \textbf{Metric} & \textbf{Description} \\
\midrule
\multirow{2}{*}{Temporal} 
& FVMD & Fréchet Video Motion Distance \cite{liu2024fr}: Focuses on motion consistency by comparing velocity and acceleration profiles of tracked keypoints; lower indicates more realistic motion patterns. \\
\midrule
\multirow{3}{*}{Spatial} 
& OC & Object-level Consistency: Cosine similarity of masked object features between adjacent frames, assessing stability of object identity and structure over time. Higher is better. \\
\cmidrule(lr){2-3}
& SC & Scene-level Consistency: masked SSIM between adjacent background regions; evaluates stability of scene lighting, texture, and geometry. Higher is better. \\
\cmidrule(lr){2-3}
& IoU & Region-level Intersection-over-Union: Compares propagated predicted masks (via SAM2 \cite{kirillov2023segment}) against reference masks for geometry/trajectory accuracy. Higher is better. \\
\midrule
Semantic 
& VLM-QA & Vision–Language Model-based physical plausibility score: Uses a physics-aware question set (penetration, gravity consistency, deformation) and maps answers to binary pass/fail; mean pass rate over all questions. The complete question set is provided in the Appendix~\ref{sec:appendix_metrics}.
 \\
\bottomrule
\end{tabular}
\label{tab:metrics}
\end{table*}

\subsection{Human Evaluation}
We conduct a human evaluation to assess the perceptual and physical quality of generated videos. Four representative models—Wan2.2~\cite{wan2.1}, CogVideo~\cite{hong2022cogvideo}, Hunyuan~\cite{kong2024hunyuanvideo}, and PhyParam (Ours)—are compared. For each model, we randomly select 30 videos (5 for each motion category), resulting in 120 videos in total. A total of 10 annotators participate in the study, ensuring that each video is independently rated by at least three annotators, yielding 360 human ratings overall. Four annotators with high-school physics backgrounds who pass a qualification test perform the final evaluation. Each annotator is shown a textual prompt and the corresponding generated video without model information and rates temporal consistency, spatial stability, and semantic alignment on a five-point Likert scale (1–5). They are instructed to treat the three aspects independently and are given example videos and clarified guidelines before beginning the task. We further correlate the automatic metrics with human preferences to verify that the benchmark is not merely an indirect proxy; the results are provided in \cref{sec:sota}. The evaluation interface, annotation examples, and further implementation details are shown in Appendix~\ref{sec:appendix_human}.

% In the parameter space, balanced sampling is performed across multiple dimensions ($e.g.$, height, direction, interaction type), effectively reducing data distribution bias and ensuring comprehensive physical coverage across diverse motion patterns.

% PhyParam-Dataset is a large-scale dataset designed for physics-consistent video generation, containing approximately 130K text--video pairs. The dataset systematically covers three categories of physical scenarios---single-body motion, two-body collision, and multi-body dynamic interaction---including representative processes such as free fall, elastic and inelastic collisions, inclined-plane motion, and externally driven dynamics. In the parameter space, balanced sampling is performed across multiple dimensions ($e.g.$, height, direction, interaction type), effectively mitigating data distribution bias and ensuring comprehensive physical coverage for diverse motion patterns.

% 虽然之前的工作已经关注到物理一致性研究的重要性，但是大多聚焦比较简单的单物体的运动方面，忽略了物体于物体以及物体于环境之间的交互，使得生成效果于自然场景之间仍然具有很大的gap。为了学习到更加复杂的物理运动，我们提出了PhyParam，其包含PhyParam-Dataset 具有详细物理参数标注的交互型物理运动数据集，同时提出了评测复杂物理运动的评估指标体系phyparam-bench。

% PhyParam包括数据集和achieves a significant expansion in multi-body interaction, continuous physical parameter control, and visual realism. It establishes a unified data foundation that integrates physical interpretability, controllable generation, and quantitative evaluation, providing a scalable, verifiable, and reproducible experimental platform for studying and benchmarking video generation models that adhere to real-world physical laws. 

%% file: sec/sec4_method.tex
\section{Method}

\begin{figure*}[t!]
    \centering
    \includegraphics[width=0.95\linewidth]{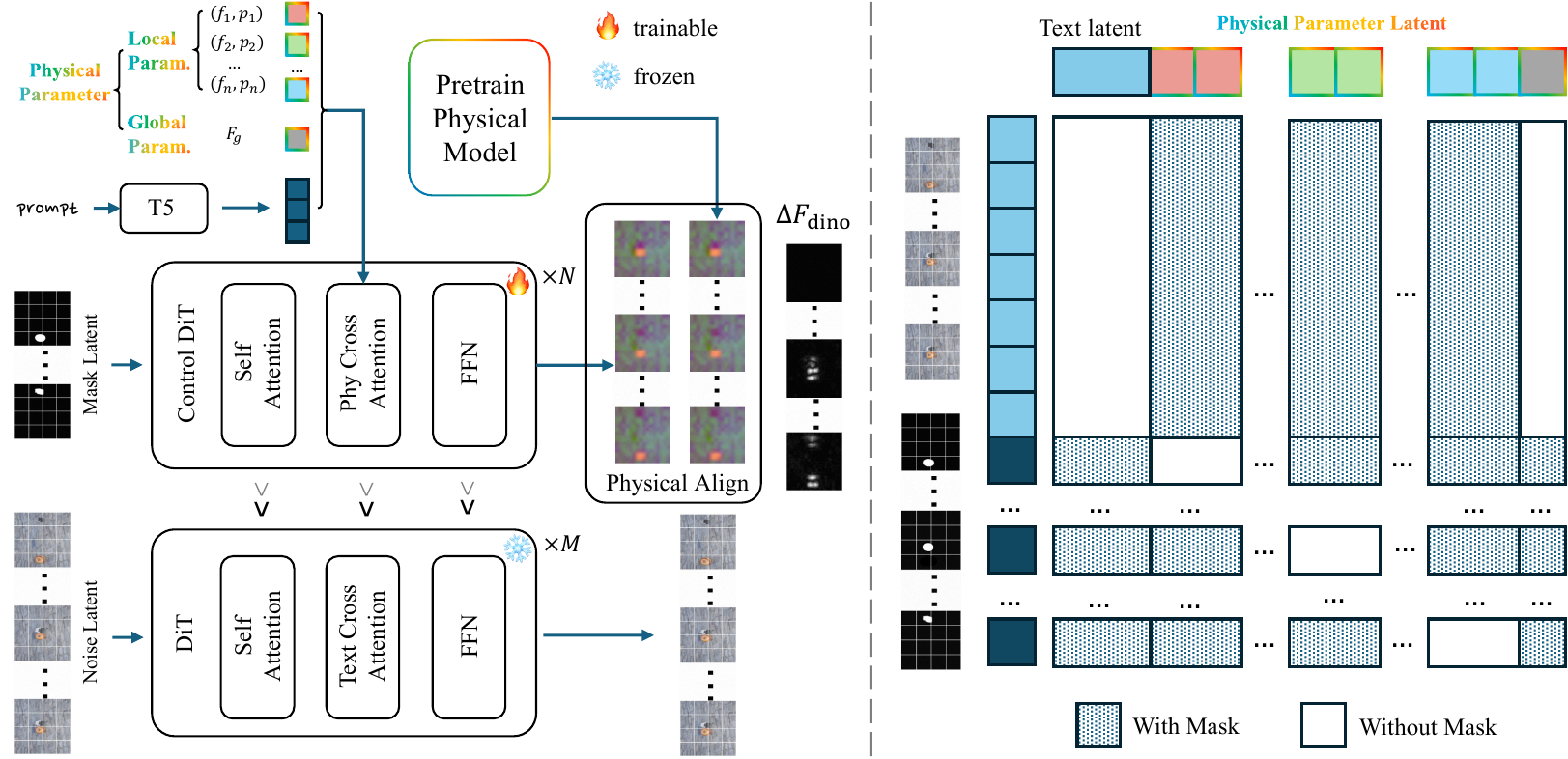}
    \caption{
Overview of the proposed physics-aware conditioning and attention.
Left: \emph{Physical Prompt} encoding. Global and local conditions are harmonic-embedded into tokens for ControlNet to avoid multi-branch overhead. Mask tokens share positional embeddings with first-frame latents for spatial alignment.
Right: \emph{Physical Cross-Attention}. Noise latents attend to mask tokens in self-attention, while cross-attention routes mask tokens to local attributes and global tokens to noise.
}
% This strengthens alignment and reduces interference while preserving base model capabilities.
\vspace{-10pt}
\label{fig:method}
\end{figure*}

\subsection{Preliminaries}
We adopt Flow Matching~\cite{lipman2022flow,esser2024scaling} to learn a continuous-time generative process for video.
During training, we sample a clean latent $x_{1}$, a noise latent $x_{0} \sim \mathcal{N}(0, I)$, and a timestep $t \in [0,1]$.
We sample $t$ from a logit-normal distribution.
We then form an intermediate latent via linear interpolation:$x_{t} = t\, x_{1} + (1 - t)\, x_{0}$, with the corresponding ground-truth velocity
\begin{equation}
v_{t} = \frac{d x_{t}}{d t} = x_{1} - x_{0}.
\label{eq:velocity}
\end{equation}

A neural network $u(x_{t}, c, t; \theta)$ is trained to predict $v_{t}$, where $c$ denotes the conditioning signal such as text.
We minimize the mean squared error:
\begin{equation}
\mathcal{L}_\textrm{flow}(\theta) =
\mathbb{E}_{x_{0}, x_{1}, c, t}
\left[
\left\lVert u(x_{t}, c, t; \theta) - v_{t} \right\rVert_{2}^{2}
\right].
\label{eq:loss}
\end{equation}

\subsection{Physical Parameter Encoding}
We build PhyParam on Wan2.2-TI2V-5B~\cite{wan2.1} and study physics-guided video generation, aiming to synthesize temporally coherent sequences whose dynamics are controlled by localized physical conditions and a shared global environment.
Concretely, each localized condition includes an external force $\mathbf{f}_i \in \mathbb{R}^{3}$, material properties $\mathbf{p}_i \in \mathbb{R}^{d_p}$, and a binary mask $\mathbf{m}_i \in \{0,1\}^{h \times w}$ indicating the target object region.
The global condition captures scene-level factors such as gravity.
In this first formulation, local forces are parameterized as center-of-mass forces for compact rigid bodies; torques induced by off-center force application are not explicitly modeled.

Existing methods often rely on text-only prompts or encode physical conditions as multi-channel control images, which makes it difficult to bind instance-level physics to the correct objects and may dilute global constraints~\cite{mao2025omni}.
While multi-branch ControlNet variants~\cite{zhang2023adding} can process different conditions independently, they introduce substantial parameter and compute overhead and may still suffer from cross-condition interference.

We represent physical conditions with a global--local decomposition:
$
\mathbf{g} \triangleq \mathbf{F}_g, 
\mathbf{c}_i \triangleq (\mathbf{m}_i, \mathbf{f}_i, \mathbf{p}_i),
$
where $\mathbf{F}_g \in \mathbb{R}^{3}$ can be a constant global field such as gravity.
Instead of instantiating multiple ControlNet branches, we serialize all conditions into a single token sequence and feed them to a unified ControlNet:
$
[
    n_{\mathrm{noise}},
    m_1,\dots,m_n,
    t_{\mathrm{text}},
    (f_1,p_1),\dots,(f_n,p_n),\, F_g 
].
$
Here, $n_{\mathrm{noise}}$ denotes the patchified noise latents.
In our implementation, each $(f_i,p_i)$ is expanded into five compact local physical-attribute tokens per object.
Each mask $m_i$ is downsampled and patchified to the latent resolution and projected to a token embedding.
All mask tokens reuse the same spatial positional embeddings as the first-frame noise latents, enabling precise spatial alignment between instance regions and diffusion features.

For the local force $\mathbf{f}_i \in \mathbb{R}^{3}$, we factorize it into direction and magnitude:
\begin{equation}
\mathbf{f}_i \mapsto \left(\hat{\mathbf{f}}_i,\ \lVert \mathbf{f}_i \rVert\right),
\qquad
\hat{\mathbf{f}}_i = \frac{\mathbf{f}_i}{\lVert \mathbf{f}_i \rVert} \in \mathbb{R}^{3}.
\end{equation}
This factorization helps the network model anisotropic effects via direction while capturing scaling behavior via magnitude.

To strengthen the representation of physical scalars, including force magnitude, direction parameters, and additional properties $\mathbf{p}_i$, we apply harmonic embedding~\cite{mildenhall2021nerf} to each attribute:
\begin{equation}
\gamma(x)=\left\{\sin\left(2^{k}\pi x\right),\ \cos\left(2^{k}\pi x\right)\ \big|\ k=0,1,\dots,L-1\right\},
\end{equation}
where $L$ is the number of frequency bands.

\subsection{Physical Parameter Attention}
To integrate multiple physical parameters without inducing cross-condition interference, we propose \emph{Physical Parameter Attention}, which injects physics signals into attention with explicit routing.
This is particularly important in multi-object or interaction-heavy scenes, where \emph{global} conditioning can easily mix instance-specific parameters.
Moreover, text prompts rarely provide precise, controllable physics, and ControlNet-style global injection may bias the model toward appearance changes while motion remains less sensitive to the intended physical values.
Extending global injection to multiple instance conditions also typically requires additional branches and non-trivial fusion design.

As shown in Fig.~\ref{fig:method}, we concatenate mask tokens with the noise latents, and concatenate physical-attribute tokens with text tokens along the sequence dimension.

In self-attention, noise latents are allowed to attend to all mask tokens, so that instance regions can be localized in the same spatiotemporal reference frame as diffusion features.
In cross-attention, to prevent condition leakage across objects, each mask token is only permitted to attend to its aligned physical-attribute tokens $(f_i, p_i)$.
In contrast, global physics tokens such as $F_g$ attend directly to the noise latents, enforcing scene-level constraints with minimal disruption to the base model.
We implement this routing using an additive attention mask or bias that blocks disallowed token interactions.

\subsection{Semantic-Structural Feature-Space Supervision}
Supervision in the VAE latent space~\cite{kingma2013auto,rombach2022high,zhao2026extending} primarily favors appearance and may under-emphasize object-level structure and temporal dynamics.
We therefore introduce a feature-space supervision scheme that operates on the ControlNet branch, aligning its condition-driven outputs with reference physical simulation results in a high-dimensional representation space.

We use a pre-trained DINOv3 model~\cite{caron2021emerging,oquab2023dinov2,simeoni2025dinov3} as a frozen feature extractor.
DINOv3 is not used to encode physical variables such as force, mass, friction, or restitution, which are injected through the parameter-conditioning branch.
Instead, it provides a semantic-structural feature space that is less sensitive to low-level appearance noise, helping align object identity, shape, boundaries, and spatial displacement with the reference simulation.

To apply this supervision, we supervise the multi-scale residual features produced by ControlNet by projecting them through an additional head to obtain feature maps $F_\textrm{ctr}$.
We extract DINO features $F_\textrm{dino}$ from the reference simulation frames and use cosine distance for feature alignment:
\begin{equation}
\mathcal{L}_\mathrm{feature} = 1 - \frac{F_\mathrm{ctr}\cdot F_\mathrm{dino}}{\|F_\mathrm{ctr}\| \ \|F_\mathrm{dino}\|}.
\end{equation}

Moreover, single-frame alignment is insufficient to capture physical dynamics.
We introduce a temporal consistency loss by matching the first-order temporal differences:
$
\Delta F_\mathrm{ctr}^{t} = F_\mathrm{ctr}^{t+1} - F_\mathrm{ctr}^{t}, 
\Delta F_\mathrm{dino}^{t} = F_\mathrm{dino}^{t+1} - F_\mathrm{dino}^{t}.
$
We again use cosine distance:
\begin{equation}
\mathcal{L}_\mathrm{temporal} = 1 - \frac{\Delta F_\mathrm{ctr}^{t}\cdot \Delta F_\mathrm{dino}^{t}}{\|\Delta F_\mathrm{ctr}^{t}\| \ \|\Delta F_\mathrm{dino}^{t}\|}.
\end{equation}

The final objective is:
$
\mathcal{L}_\mathrm{total} = \mathcal{L}_\mathrm{flow} + \lambda_1 \mathcal{L}_\mathrm{feature} + \lambda_2 \mathcal{L}_\mathrm{temporal},
$
where $\lambda_1$ and $\lambda_2$ balance feature alignment and temporal consistency, respectively.

%% file: sec/sec5_experiment.tex
\section{Experiment}

\subsection{Experimental Settings}

\noindent\textbf{Implementation Details.} We run all experiments on a cluster with 8 NVIDIA A100 GPUs. For each run and each experimental setting, we train the model for 5{,}000 optimization steps with a learning rate of $1\times 10^{-4}$. We use a per-GPU batch size of 1 and accumulate gradients for 4 steps, resulting in an effective batch size of 32. All experiments use mixed-precision training with \texttt{bfloat16} to improve throughput and reduce memory usage.

\noindent\textbf{Baseline.} We build our method on top of Wan2.2-TI2V-5B~\cite{wan2.1}, a recent state-of-the-art video diffusion model. For comparison, we evaluate representative video generation baselines, including HunyuanI2V-14B~\cite{kong2024hunyuanvideo}, CogVideoX-5B~\cite{hong2022cogvideo}, and the physics-guided Force Prompting~\cite{gillman2025force}.

\noindent\textbf{Fairness of Evaluation.}
We standardize conditioning and avoid privileged inputs to ensure a fair comparison across models with different control interfaces.
For each PhyParam-Bench instance, we provide a fixed condition package (caption, physical parameters, and first-frame instance masks exported from Blender).
Each baseline only receives inputs supported by its native interface while sharing the same first-frame image and evaluation protocol: text-only models are prompted with the physical information, ControlNet-style models take text and mask, and models with explicit physics-conditioning take parameters through their dedicated channels.

% \noindent\textbf{Baseline.} We compare our method against two categories of video generation models. The first category consists of general image-to-video models, including Wan2.2-5B, HunyuanI2V-14B, and CogVideoX-5B. For these methods, we provide the mechanical parameters in textual form along with the initial frame image as input, prompting the model to generate the subsequent motion frames. The second category focuses on publicly available physics-consistent video generation models, represented by Force-Prompting. For a fair comparison, we select a subset from our benchmark—specifically, the single-object horizontal motion test set—and convert our mechanical conditions into the input format required by Force-Prompting. We also supply the first-frame mask to this model and evaluate its generated motion trajectories.

\begin{table}[htbp]
\centering
\scriptsize
\caption{\textbf{Physics-related evaluation on PhyParam-Bench.}
Best results are \textbf{bolded} within each group (General vs. Physics-guided)}
\label{tab:phy_metrics}
\renewcommand{\arraystretch}{1.15}
\setlength{\tabcolsep}{3pt}
\resizebox{0.75\linewidth}{!}{%
\begin{tabular}{
    ll
    S[round-mode=places,round-precision=1]
    S[round-mode=places,round-precision=1]
    S[round-mode=places,round-precision=1]
    S[round-mode=places,round-precision=1]
    S[round-mode=places,round-precision=1]
}
\toprule
\multicolumn{2}{c}{\textbf{Method}} &
{\textbf{FVMD$\downarrow$}} &
{\textbf{IoU$\uparrow$}} &
{\textbf{OC$\uparrow$}} &
{\textbf{SC$\uparrow$}} &
{\textbf{VLM-QA$\uparrow$}} \\
\midrule
\multirow{4}{*}{\textbf{General Models}}
 & CogVideoX        & \num{1210.7} & \num{13.0} & \num{52.9} & \num{93.3} & \num{62.3} \\
 & HunyuanI2V       & \num{636.9}  & \num{15.2} & \num{54.9} & \num{96.3} & \num{66.5} \\
 & Wan2.2-TI2V-5B   & \num{1141.8} & \num{11.4} & \num{54.9} & \num{86.6} & \num{56.3} \\
 & Ours-All         & \textbf{\num{355.5}} & \textbf{\num{17.8}} & \textbf{\num{58.1}} & \textbf{\num{96.4}} & \textbf{\num{69.5}} \\
\midrule
\multirow{2}{*}{\textbf{Physical Models}}
 & Force Prompting  & \textbf{\num{455.3}} & \num{6.2} & \textbf{\num{74.3}} & \textbf{\num{97.1}} & \num{74.0} \\
 & Ours-Force       & \num{586.8} & \textbf{\num{8.4}} & \num{54.4} & \num{96.3} & \textbf{\num{74.4}} \\
\bottomrule
\end{tabular}%
}
\end{table}

\subsection{Comparison with State-of-the-Art}
\label{sec:sota}

\begin{figure*}[t]
    \centering
    \includegraphics[width=0.95\linewidth]{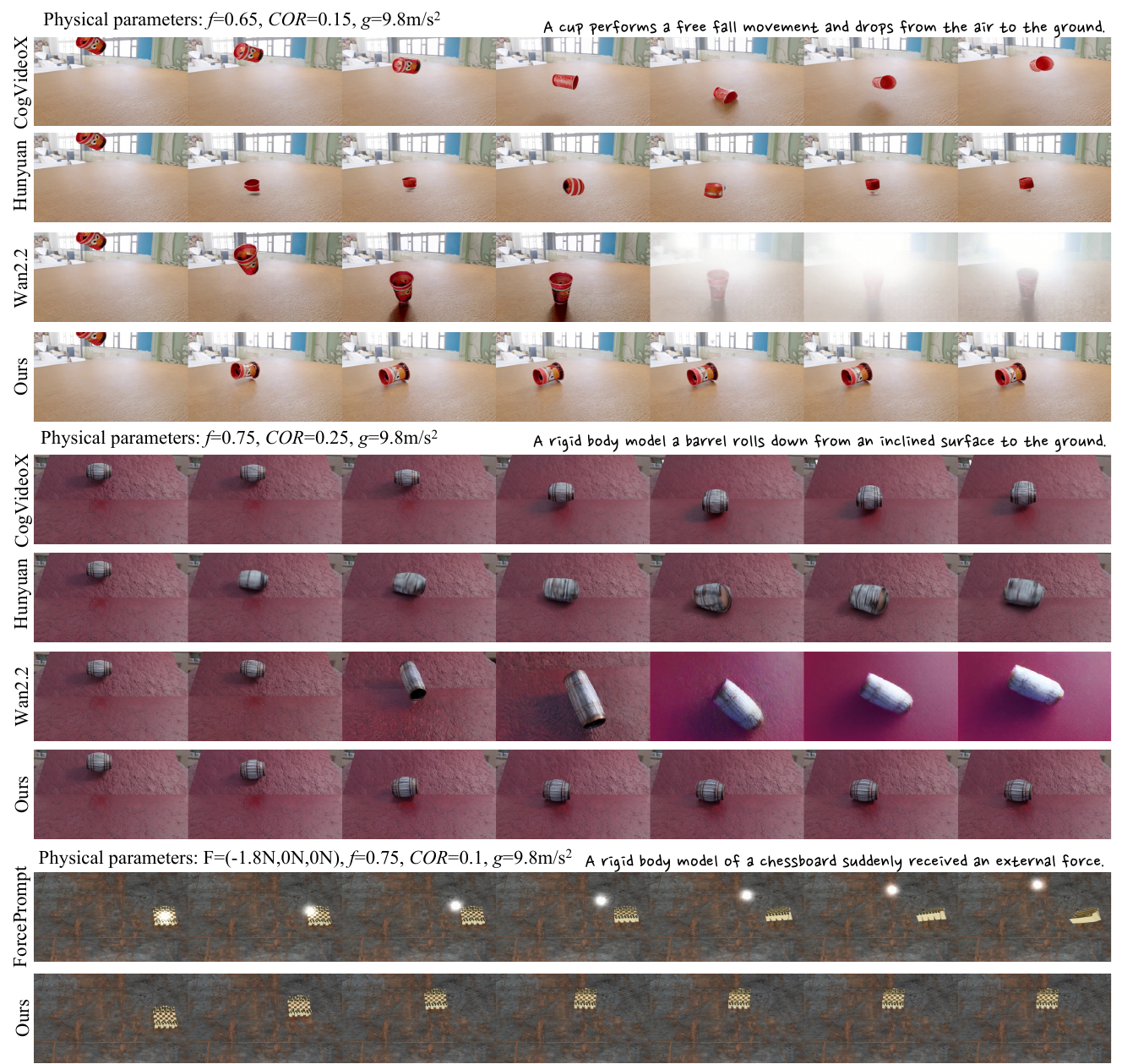}
    
    \caption{Qualitative Comparison. We compare our results with Wan2.2-TI2V-5B~\cite{wan2.1},
CogVideoX-5B~\cite{hong2022cogvideo}, and HunyuanI2V-14B~\cite{kong2024hunyuanvideo}.
We further compare against Force Prompting~\cite{gillman2025force} for object motion on horizontal surfaces
(The moving light point visualizes the input conditioning provided to the model).}
    \label{fig:con}
\end{figure*}

\noindent\textbf{Quantitative Comparison on PhyParam-Bench.}
We benchmark recent image-to-video models on PhyParam-Bench using the metrics in \cref{tab:phy_metrics}.
Overall, \textbf{Ours-All} delivers the most \emph{balanced} performance across perceptual quality and physical consistency, achieving the best FVMD (355.5) and consistently higher scores on interaction- and stability-related metrics (IoU 17.8, OC 58.1, SC 96.4).
This indicates that incorporating explicit physical signals improves both temporal coherence and spatially stable interactions.

Among general-purpose baselines, CogVideoX~\cite{hong2022cogvideo} and Wan2.2-TI2V-5B~\cite{wan2.1} exhibit larger distortion and weaker temporal alignment, while HunyuanI2V-14B~\cite{kong2024hunyuanvideo} is comparatively stronger yet still struggles with stable physical interactions under our long-tail settings.
Among physics-guided approaches, \textbf{Ours-Force} is competitive with Force Prompting~\cite{gillman2025force}, achieving higher semantic alignment (VLM-QA 74.4) with comparable physical coherence.
While Force Prompting attains a lower FVMD, it yields lower IoU, suggesting limited object-level controllability in collision- and contact-heavy scenarios.
Taken together, these results support that our \textbf{Physical Parameter Attention} improves controllability and physical plausibility without sacrificing visual quality.

\noindent\textbf{Human Evaluation.}
We perform a human evaluation on 120 cases to assess perceived \emph{temporal consistency}, \emph{spatial consistency}, and \emph{semantic consistency} (\cref{tab:human_evaluation}).
Our method achieves the highest preference rates on all three dimensions (74.48\%, 68.97\%, and 75.17\%, respectively), indicating stronger human-perceived physical plausibility and overall coherence.
Compared with CogVideoX, HunyuanI2V-14B, and Wan2.2-TI2V-5B, our outputs are preferred for smoother trajectories, reduced object interpenetration, and more consistent motion--context alignment.
These observations support the effectiveness of explicit physical-parameter conditioning in improving both visual fidelity and perceived physical realism.

We further validate the faithfulness of PhyParam-Bench by correlating its metrics with human judgments.
Across dimensions, we observe a strong Spearman rank correlation ($\rho = 0.80$), suggesting that the proposed metrics serve as reliable proxies for human-perceived physical consistency.
Finally, we report VBench results to evaluate general video quality beyond physics-specific factors, with additional details in Appendix~\ref{sec:appendix_exper}.

% \begin{table*}[htbp]
% \centering
% \small
% \caption{
% Performance comparison on Physics-related Metrics (\emph{Phy-Metrics}). 
% Best value in each column is \textbf{bolded} separately for General and Physical models.
% }
% \label{tab:phy_metrics}
% \renewcommand{\arraystretch}{1.15}
% \setlength{\tabcolsep}{5pt}
% \begin{tabular}{
%     l % Method names
%     S[table-format=6.2] % FVMD
%     S[table-format=1.2] % IoU
%     S[table-format=1.2] % OC
%     S[table-format=1.2] % SC
%     S[table-format=2.2] % VLM-QA
% }
% \toprule
% \textbf{Method} &
% % {\textbf{FVD}$\downarrow$} &
% {\textbf{FVMD}$\downarrow$} &
% {\textbf{IoU}$\uparrow$} &
% {\textbf{OC}$\uparrow$} &
% {\textbf{SC}$\uparrow$} &
% {\textbf{VLM-QA}$\uparrow$} \\
% \midrule
% \multicolumn{6}{l}{\textbf{General Models}} \\
% CogVideoX & 1210.70 &  946.61 & 13.02 & 52.94 & 93.27 & 62.28 \\
% Hunyuan   &  636.85 & 1259.92 & 15.17 & 54.88 & 96.27 & 66.53 \\
% Wan~2.2   & 1141.83 & 28036.00 & 11.40 & 54.94 & 86.64 & 56.26 \\
% Ours-All  & \textbf{355.48} & \textbf{17.77} & \textbf{58.11} & \textbf{96.38} & \textbf{69.46} \\
% \midrule
% \multicolumn{6}{l}{\textbf{Physical Models}} \\
% Force Prompting & \textbf{455.29} & \textbf{219.98} & 6.23 & \textbf{74.31} & \textbf{97.12} & \textbf{0.74} \\
% Ours-Force      & 586.82 & 879.63 & \textbf{8.37} & 54.41 & 96.31 & 74.43 \\
% \bottomrule
% \end{tabular}
% \end{table*}

\begin{table}[t]
\centering
\scriptsize
\caption{Human evaluation results on temporal, spatial, and semantic consistency. Higher is better.}
\label{tab:human_evaluation}
\resizebox{0.55\linewidth}{!}{%
\begin{tabular}{lccc}
\toprule
\textbf{Method} & \textbf{Temporal}~$\uparrow$ & \textbf{Spatial}~$\uparrow$ & \textbf{Semantic}~$\uparrow$ \\
\midrule
CogVideoX-5B        & 34.48 & 20.69 & 31.03 \\
HunyuanI2V-14B      & 46.90 & 37.93 & 48.97 \\
Wan2.2-TI2V-5B      & 37.93 & 40.69 & 35.86 \\
Ours                & \textbf{74.48} & \textbf{68.97} & \textbf{75.17} \\
\bottomrule
\end{tabular}%
}
\end{table}

\subsection{Ablation Study}

\begin{table}[t]
\centering
\scriptsize
\caption{Ablation on loss design. \textbf{Feature alignment} denotes the DINO-based semantic-structural feature alignment loss, and \textbf{Diff feature alignment} denotes the temporal diffusion feature alignment loss.}
\label{tab:phy_metrics_modules}
\renewcommand{\arraystretch}{1.1}
\setlength{\tabcolsep}{4pt}
\resizebox{0.85\linewidth}{!}{
\begin{tabular}{ccccccc}
\toprule
\textbf{Feature align} & \textbf{Diff feature align} &
\textbf{FVMD$\downarrow$} & \textbf{IoU$\uparrow$} & \textbf{OC$\uparrow$} & \textbf{SC$\uparrow$} & \textbf{VLM-QA$\uparrow$} \\
\midrule
\xmark & \xmark & 801.9 & 16.2 & 57.7 & 96.1 & 68.5 \\
\cmark & \xmark & 408.4 & 17.3 & 58.1 & 96.3 & 69.0 \\
\cmark & \cmark & \textbf{355.5} & \textbf{17.8} & \textbf{58.4} & \textbf{96.4} & \textbf{69.5} \\
\bottomrule
\end{tabular}}
\end{table}

\begin{table}[htbp]
    \centering
    \scriptsize
    \vspace{-1.0em}
    \caption{Ablation on feature representation and Physical Attention (Phy-Att).}
    %  \textbf{Type} indicates the backbone used for feature-space supervision (DINOv3 or VJEPAv2). \textbf{Phy-Att} denotes Physical Attention for instance-level binding.
    \tiny
    \footnotesize
    \resizebox{0.60\linewidth}{!}{
    \begin{tabular}{lcccccc}
    \toprule
    \textbf{Type} & \textbf{Phy-Att} &\textbf{FVMD$\downarrow$} & \textbf{IoU$\uparrow$} & \textbf{OC$\uparrow$} & \textbf{SC$\uparrow$} & \textbf{VLM-QA$\uparrow$}\\
    \midrule
    Dinov3 & \cmark & \textbf{355.5} & \textbf{17.8} & \textbf{58.1} & \textbf{96.4} & \textbf{69.5} \\
    Dinov3 & \xmark & 473.4 & 11.9 &56.2 &96.2 &68.9 \\
    VJEPAv2 & \cmark & 422.9 & 16.5 &57.8 &96.0 &68.3 \\
    \bottomrule
    \end{tabular}}
    \label{tab:abl_result}
    \vspace{-1.5em}
\end{table}

% /CVPR Physical/fig/dino.pdf

\begin{figure}[t]
    \centering
    \includegraphics[width=0.7\linewidth]{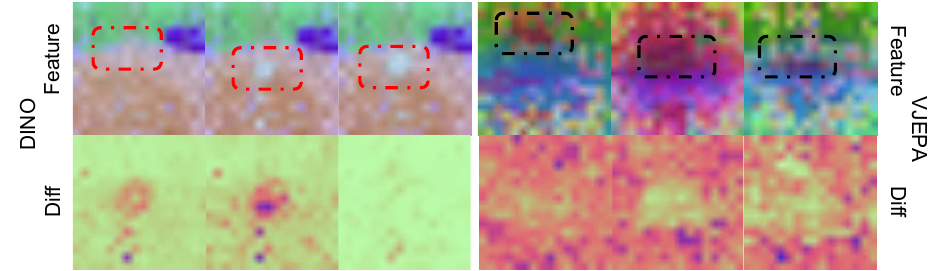}
    \caption{Compared with VJEPA, DINO features are temporally more stable, making them better suited for semantic-structural feature-space supervision.}
    \label{fig:dino}
\end{figure}

\noindent\textbf{Effect of Loss Configuration.} We study how the loss design affects physical consistency on PhyParam-Bench (\cref{tab:phy_metrics_modules}). Without either loss, the model exhibits weaker temporal coherence and less stable object interactions, reflected by a much higher FVMD. Adding only the DINO-based semantic-structural feature alignment substantially improves FVMD and consistently boosts IoU and VLM-QA, suggesting that aligning high-level representations helps preserve object semantics and spatial structure. When both feature alignment and temporal diffusion feature alignment are enabled, the model achieves the best overall performance, with further gains in FVMD, IoU, SC, and VLM-QA. 
In particular, FVMD decreases from 801.9 without feature-space supervision to 408.4 with semantic-structural feature alignment, and further to 355.5 after adding temporal feature alignment.

% As visualized in \cref{fig:dino}, the two losses are complementary: feature alignment improves spatially meaningful activations, while diffusion feature alignment stabilizes temporal dynamics, leading to more coherent motion and interactions. Additional ablations are provided in Appendix~\ref{sec:appendix_exper}.

\noindent\textbf{Physical Attention for instance-level physics binding.}
A key challenge in multi-object and interaction-heavy scenes is to bind instance-specific physical parameters to the correct objects.
Na\"ive global conditioning can mix attributes across instances, making motion less sensitive to the intended physics even when appearance changes.
Our Physical Attention addresses this by explicitly attending to target instances and injecting the corresponding physical attributes, improving the alignment from \emph{physics conditions} to \emph{resulting motion}.
As shown in \cref{tab:abl_result}, enabling Physical Attention consistently improves physical-consistency metrics, especially those reflecting interaction stability.
Removing it worsens FVMD from 355.5 to 473.4 and IoU from 17.8 to 11.9, confirming that object-level routing is critical for reducing cross-instance condition leakage.

\noindent\textbf{Physics-aware feature-space supervision.}
Standard diffusion training in VAE latent space prioritizes appearance reconstruction and may under-emphasize object-level structure and temporal consistency. We therefore introduce feature-space supervision using DINO features. DINO does not directly represent physical variables; rather, it supplies a semantic-structural space for aligning object identity, shape, boundaries, and spatial displacement with the reference simulation while the explicit physical parameters are injected by the conditioning branch. We also evaluated alternative representations, including motion-centric features, and observed that VJEPA features are noise-sensitive and temporally unstable. This instability persists across frames and is evident even in regions that remain static, which results in noisier supervision and slower convergence. In contrast, DINO offers a better trade-off among training stability, physics consistency, and visual quality, as supported by the ablations in \cref{tab:abl_result} and the feature visualizations in \cref{fig:dino}. It shows that VJEPA exhibits pronounced temporal fluctuations, whereas DINO remains substantially more stable.

\subsection{Visualization Analysis}

\noindent\textbf{Qualitative evaluation on physical interactions.}
We qualitatively compare PhyParam with representative baselines to assess physical consistency (\cref{fig:con}). HunyuanI2V-14B generates visually appealing videos but often violates basic physics, e.g., floating objects or unrealistic sliding. CogVideoX-5B shows better temporal continuity but struggles to preserve rigid-body properties. Wan2.2-TI2V-5B produces smoother transitions yet tends to oversimplify contact dynamics, while Force Prompting introduces limited local corrections but may destabilize global motion. In contrast, PhyParam yields more stable and physically plausible interactions with better shape preservation and momentum transfer, consistent with the quantitative results in \cref{tab:phy_metrics}. Failure cases are discussed in Appendix~\ref{sec:appendix_exper}.

\begin{figure}[t]
  \centering
  \includegraphics[width=0.9\linewidth]{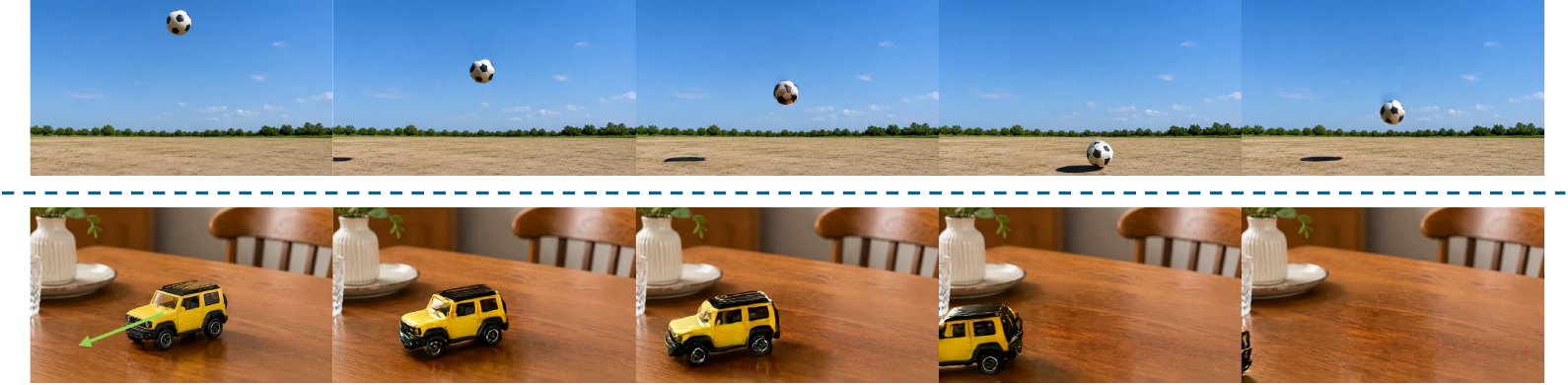}
  \vspace{2pt}
  \includegraphics[width=0.9\linewidth]{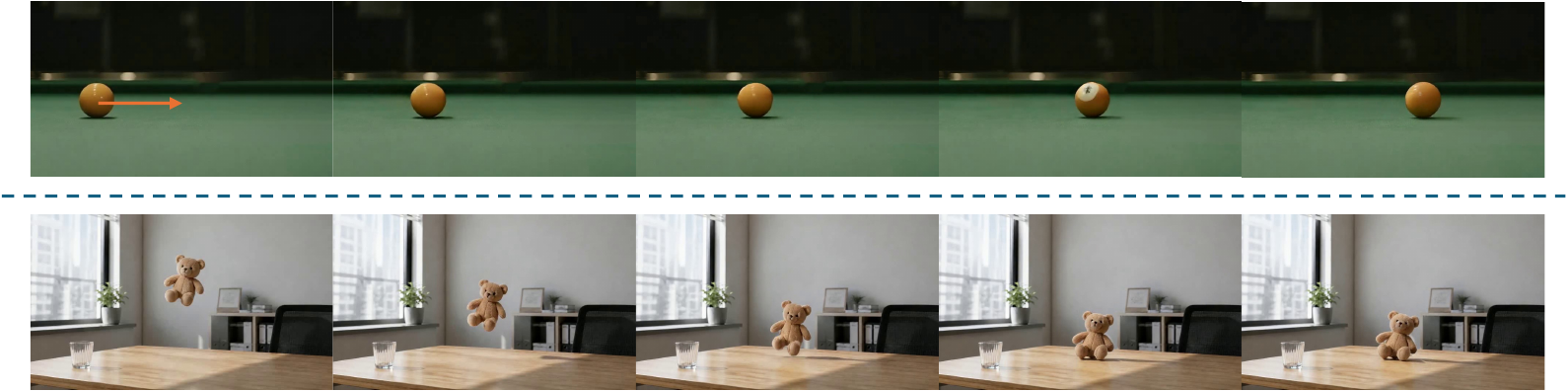}
  \caption{Additional qualitative validation. Top: real-image inputs with manually annotated masks and specified physical parameters. Bottom: generated dynamics under varying physical parameters while keeping the initial image fixed.}
   \label{fig:real_param_cases}
\end{figure}

\begin{figure}[h]
  \centering
  \vspace{-1.0em}
  \includegraphics[width=0.9\linewidth]{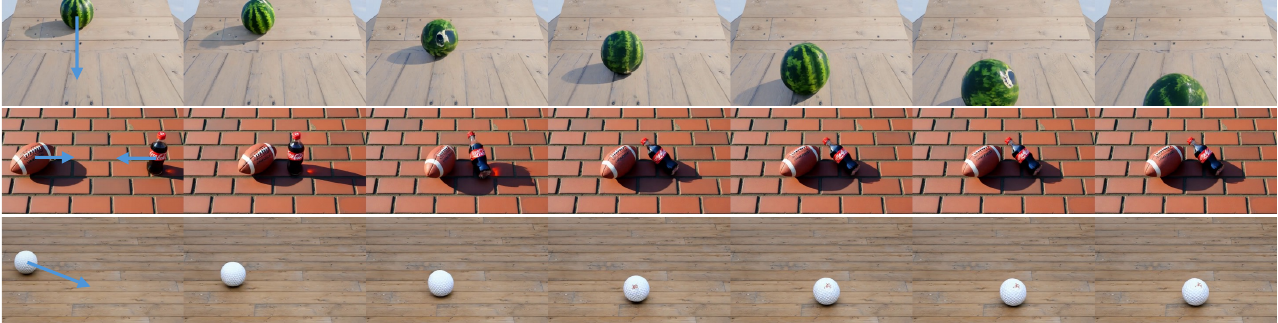}
  \caption{Qualitative I2V results on real-world-like videos covering diverse motion patterns. (Arrows indicate object motion directions.) }
   \label{fig:real}
\end{figure}

\noindent\textbf{Real-image validation and parameter sensitivity.}
Synthetic simulation provides controllable and verifiable supervision, including physical parameters, trajectories, and instance masks that are difficult to obtain at scale in real scenes.
Beyond the synthetic benchmark, we test PhyParam on real images with manually annotated masks and specified physical parameters.
As shown in \cref{fig:real_param_cases}, the model produces plausible rigid-body motion for real-image inputs, suggesting preliminary synthetic-to-real transfer under the same I2V interface.
We also vary physical parameters, including force, friction, and restitution, while keeping the initial image and target masks fixed, and observe corresponding changes in generated dynamics, indicating that the model responds to force and material-property controls rather than relying only on appearance priors.
For T2I applications, a text-to-image model such as FLUX can first generate the initial frame, after which masks are obtained manually or by segmentation and PhyParam synthesizes the physics-conditioned video.
To better cover the motion types studied in this work, we also provide additional ``real-world-like'' cases in \cref{fig:real}, where the evaluation images are generated using FLUX with manually designed prompts.
\vspace{-20pt}

%% file: sec/sec6_conclusion.tex
\section{Conclusion}

We introduced \textbf{PhyParam}, a unified framework for explicit rigid-body physical-parameter control in image-to-video generation.
Our work comprises \textbf{PhyParam-Dataset}, a large-scale corpus of physically simulated videos with explicit parameter annotations; \textbf{PhyParam-Bench}, a multi-level benchmark for assessing physical-law consistency; and the \textbf{PhyParam} model, which injects object-level and scene-level physical parameters into the generative process to produce more controllable and interpretable motion.

\noindent\textbf{Limitations.}
The current formulation focuses on compact rigid bodies and assumes local forces are applied at or near the object center of mass.
It does not explicitly model torque from off-center force application; it also does not cover deformable objects, fluids, cloth, fracture, soft bodies, or articulated dynamics.
Looking ahead, we plan to couple generative models with differentiable physics and extend our formulation to richer interaction types.

%% file: sec/X_suppl.tex
\clearpage
\setcounter{section}{0}       % 章节编号从 1 开始计数
\renewcommand{\thesection}{S\arabic{section}} % 可选：附录章节编号如 S1, S2...

\section{Dataset Generation Details}
\label{sec:appendix_dataset}
\begin{figure}[htbp]
    \centering
    \includegraphics[width=\linewidth]{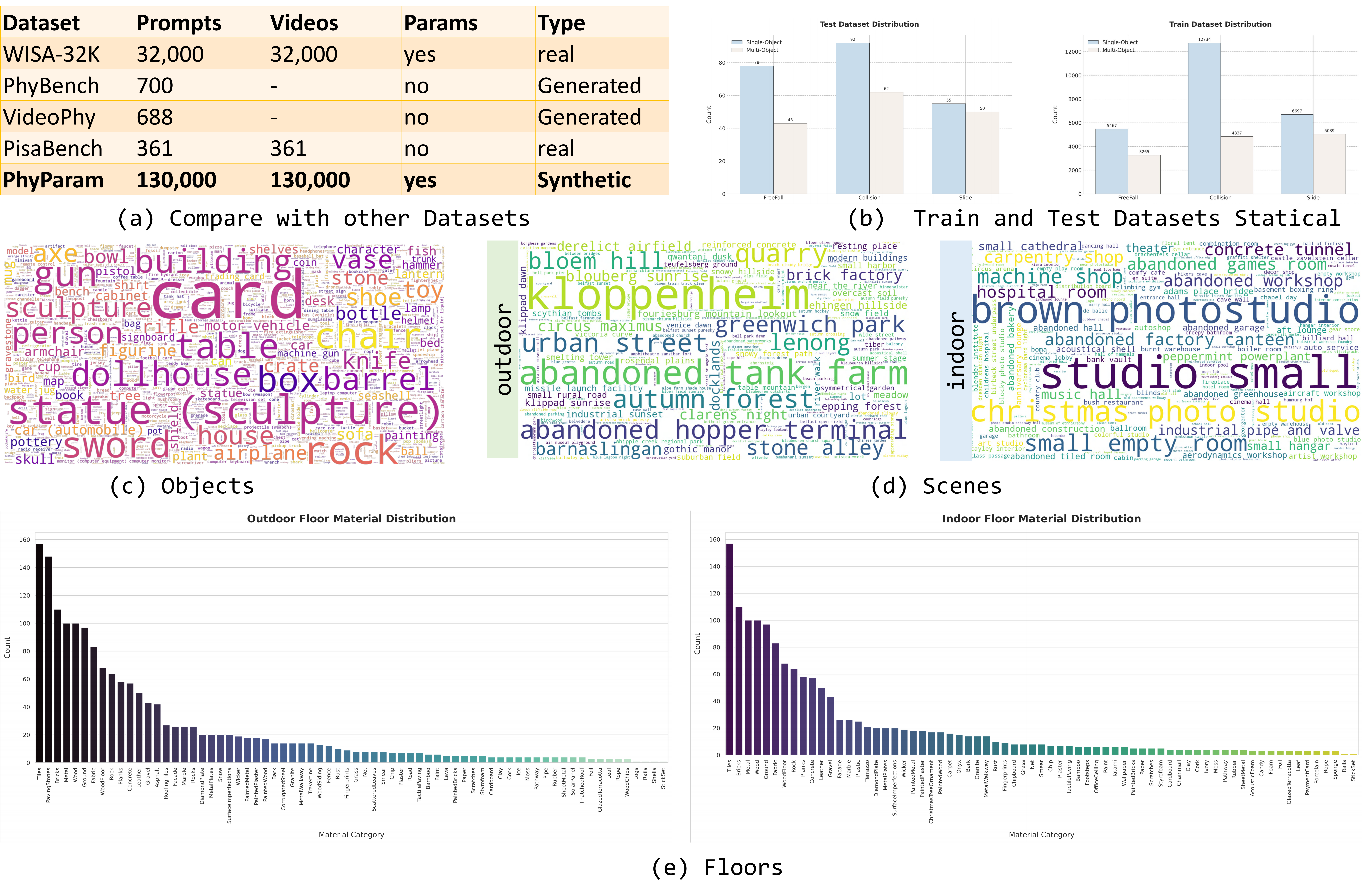} % 
    \caption{
        Distribution of CC-textures and PolyHaven environment maps between indoor and outdoor categories. 
    }
    \label{fig:texture_categories}
\end{figure}

\noindent\textbf{Basic Scene Setup.}
We start by placing a sufficiently large planar surface at the scene center, which serves as the support for all physical interactions.
We then add a directional light source tilted at $45^\circ$ relative to the plane, along with a camera for recording the rendered videos.
To increase dataset diversity, we randomly sample the camera pose from five predefined viewpoints.
After establishing this basic layout, we incorporate CC-textures for planar surfaces and PolyHaven environment maps.
These assets are automatically categorized as indoor or outdoor using a large-scale vision--language model, and are applied to the corresponding scene settings to improve visual realism and appearance diversity.
Figure~\ref{fig:texture_categories} shows the distribution of the materials across these categories.

\noindent\textbf{Placement of moving subjects.}
Moving subjects are sampled from 64K high-quality annotated 3D models in Objaverse.
For each model, we load it into the scene, scale it to fit entirely within the camera frustum, and translate it to its designated location.
We define two main categories of physical interaction:
(i) single-object interactions with the environment, and
(ii) multi-object interactions, including both multi-object--environment and object--object scenarios.
In single-object setups, placement locations are randomly sampled within a predefined range.
In multi-object setups, we compute the bounding boxes of existing objects and place each newly added object outside these bounds to prevent mesh interpenetration.
Within each interaction category, we consider three motion types:
(i) \emph{free-fall}, where objects are initialized at random heights within the camera frustum;
(ii) \emph{collision}, where objects are placed on a plane with perfect surface contact; and
(iii) \emph{slide}, where we add an inclined plane at $45^\circ$ and position objects according to their bounding-box dimensions.
The full placement procedure is summarized in Algorithm~\ref{alg:placement}.

\begin{algorithm}[htbp]
\caption{Placement ensuring visibility and physical plausibility across motion types.}
\label{alg:placement}
\KwIn{Number of scenes N\_scenes, camera parameters, Objaverse dataset}
\KwOut{Scene layouts with placed objects and motion type labels}

\For{scene\_id $\gets$ 1 \KwTo N\_scenes}{
    interaction\_type $\gets$ random choice from \{single-object, multi-object\} \;

    \eIf{interaction\_type is single-object}{
        num\_objects $\gets$ 1 \;
    }{
        num\_objects $\gets$ 2 \;
    }

    objects $\gets$ empty list \;

    \For{i $\gets$ 1 \KwTo num\_objects}{
        obj $\gets$ load\_random\_objaverse\_model() \;
        obj.set\_position(0, 0, 0) \;

        S $\gets$ calc\_scale\_matrix\_to\_fit\_camera(obj, camera) \;
        obj.apply\_scale(S) \;

        \eIf{interaction\_type is multi-object \textbf{and} $i > 1$}{
            pos $\gets$ find\_position\_outside\_bboxes(obj.bbox, objects.bboxes) \;
        }{
            pos $\gets$ random\_position\_within\_range() \;
        }
        T $\gets$ calc\_translation\_matrix(pos) \;
        obj.apply\_translation(T) \;

        append obj to objects \;
    }

    motion\_type $\gets$ random choice from \{free-fall, collision, slide\} \;

    \uIf{motion\_type is free-fall}{
        \ForEach{obj $\in$ objects}{
            h $\gets$ random\_uniform(h\_min, h\_max) \;
            obj.set\_position(z = h, within\_camera = True) \;
        }
    }
    \uElseIf{motion\_type is collision}{
        \ForEach{obj $\in$ objects}{
            plane\_pos $\gets$ random\_position\_on\_plane(contact = True) \;
            obj.set\_position(plane\_pos) \;
        }
    }
    \uElseIf{motion\_type is slide}{
        plane $\gets$ add\_inclined\_plane(angle = 45$^\circ$) \;
        \ForEach{obj $\in$ objects}{
            pos $\gets$ calc\_position\_on\_incline(obj.bbox, plane) \;
            obj.set\_position(pos) \;
        }
    }

    save\_scene\_layout(scene\_id, objects, interaction\_type, motion\_type) \;
}
\end{algorithm}

\noindent\textbf{Realistic physical simulation.}
We employ a fine-grained, physics-based simulation to capture realistic object dynamics.
After placement, each object is configured as a rigid body and assigned fundamental physical properties, including a restitution coefficient (0--1), a friction coefficient (0--1), mass, and density.
We also define a global gravity constant for the entire scene.
The simulation strategy is conditioned on the motion type.
For \emph{free-fall} and \emph{slide}, objects are placed at their target locations in the first frame and then evolve freely under Blender's built-in physics engine.
For \emph{collision} on a horizontal plane, objects are keyframed in both the first and tenth frames to emulate an initial external impulse; starting from frame 10, Blender's physics simulation runs without additional control inputs.
The full pseudocode is provided in Algorithm~\ref{alg:simulation} in the supplementary material.

\begin{algorithm}[!htb]
\caption{Simulate rigid bodies under physical forces according to motion type.}
\label{alg:simulation}
\KwIn{Scene layout with placed objects and motion type}
\KwOut{Simulated trajectories and rendered outputs}

\ForEach{scene}{
    load objects, motion\_type \;

    \ForEach{obj $\in$ objects}{
        obj.restitution $\gets$ random in [0, 1] \;
        obj.friction $\gets$ random in [0, 1] \;
        obj.mass, obj.density $\gets$ random physical values \;
        set\_rigid\_body(obj) \;
    }

    apply\_global\_gravity($G$) \;

    \uIf{motion\_type is free-fall}{
        place\_in\_target\_position(objects, frame1) \;
        run\_blender\_physics(objects, mode = gravity\_only) \;
    }
    \uElseIf{motion\_type is collision}{
        set\_position(objects, frame1, plane\_start) \;
        set\_position(objects, frame10, plane\_target) \;
        run\_blender\_physics(objects, start\_frame = 10) \;
    }
    \uElseIf{motion\_type is slide}{
        place\_on\_inclined\_plane(objects, angle = 45$^\circ$) \;
        run\_blender\_physics(objects, mode = gravity\_with\_friction) \;
    }

    save\_simulation\_data(scene) \;
}
\end{algorithm}

\noindent\textbf{Video rendering.}
After scene construction and physical simulation, we render each sequence using Blender's Cycles engine with 1{,}024 samples, a resolution of $480\times720$ pixels, and a frame rate of 24\,fps.
We export multiple modalities, including RGB frames, depth maps, surface normals, and semantic segmentation labels.
All rendered outputs, together with the corresponding configuration parameters, are stored in HDF5 format for efficient loading and downstream processing.

\section{More Details about Metrics}
\label{sec:appendix_metrics}

\noindent\textbf{Notation.}
Let a generated video be $V=\{I_t\}_{t=1}^{T}$, where $I_t$ denotes the RGB frame at time $t$ over the pixel domain $\Omega$ (with pixels indexed by $p\in\Omega$).
If a paired reference is available, we denote it by $\tilde V=\{\tilde I_t\}_{t=1}^{T}$.
For foreground--background separation, let $S_t\in\{0,1\}^{|\Omega|}$ be the subject (foreground) mask and $B_t=1-S_t$ the background mask.
For region-level evaluation, $M_t$ and $\hat M_t$ denote the reference and predicted region masks at frame $t$, respectively.
We use a frozen image encoder $\phi(\cdot)$ (e.g., ViT/CLIP) to extract features when needed.
Unless otherwise noted, all averages are taken over the corresponding index ranges.

\noindent\textbf{Temporal Dynamics.} 

% \noindent\textit{Fr\'echet Video Distance (FVD):} We extract clip-level features with a fixed I3D (or equivalent) and fit Gaussians to the reference and generated feature sets:
% $\mathcal{N}(\mu_r,\Sigma_r)$ and $\mathcal{N}(\mu_g,\Sigma_g)$, where $\mu_\star$ and $\Sigma_\star$ are the empirical mean and covariance in feature space.
% The FVD is the Fr\'echet distance between these Gaussians:
% \begin{equation}
% \label{eq:fvd}
% \mathrm{FVD}
% = \|\mu_r-\mu_g\|_2^2
% + \mathrm{Tr}\!\big(\Sigma_r+\Sigma_g
% -2(\Sigma_r^{\frac12}\Sigma_g\,\Sigma_r^{\frac12})^{\frac12}\big).
% \end{equation}
% Lower is better, indicating closer global dynamics to the reference distribution.

% \noindent\textit{Fr\'echet Video Distance (FVD):} We extract clip-level features with a fixed I3D (or equivalent) and fit Gaussians to the reference and generated feature sets:
% $\mathcal{N}(\mu_r,\Sigma_r)$ and $\mathcal{N}(\mu_g,\Sigma_g)$, where $\mu_\star$ and $\Sigma_\star$ are the empirical mean and covariance in feature space.
% The FVD is the Fr\'echet distance between these Gaussians:
% \begin{equation}
% \label{eq:fvd}
% \mathrm{FVD}
% = \|\mu_r-\mu_g\|_2^2
% + \mathrm{Tr}\!\big(\Sigma_r+\Sigma_g
% -2(\Sigma_r^{\frac12}\Sigma_g\,\Sigma_r^{\frac12})^{\frac12}\big).
% \end{equation}
% Lower is better, indicating closer global dynamics to the reference distribution.
\noindent\textit{(a) Fr\'echet Video Distance}~\textup{(FVD):}We use the Fréchet Video Distance to assess the overall quality of generated videos in terms of both spatial appearance and temporal dynamics. Clip-level spatiotemporal features are extracted using a fixed I3D network (or equivalent), and the distributions of reference and generated features are modeled as multivariate Gaussians characterized by their empirical means and covariances in the feature space. Lower scores indicate closer global dynamics to the reference distribution.

\noindent\textit{(b) Fr\'echet Video Motion Distance} ~\textup{(FVMD):}While FVD captures holistic visual and temporal fidelity, it is less sensitive to fine-grained motion consistency. We adopt the Fréchet Video Motion Distance as a complementary metric focusing on motion temporal consistency. Keypoint motion trajectories are extracted using the pre-trained point tracking model PIPs++, from which velocity and acceleration profiles are computed. Since realistic physical motions rarely exhibit abrupt changes in acceleration, FVMD evaluates the statistical properties of these motion vectors and compares the motion features of generated and reference videos via the Fréchet Distance, with lower scores indicating more consistent motion patterns.

\begin{table*}[t]
\centering
\caption{Performance comparison on VBench metrics.
Best values in each column are \textbf{bolded}.}
\label{tab:vbench_metrics}

\resizebox{\linewidth}{!}{%
\begin{tabular}{
    ll
    S[table-format=2.2]
    S[table-format=2.2]
    S[table-format=2.2]
    S[table-format=2.2]
    S[table-format=2.2]
    S[table-format=2.2]
}
\toprule
\multicolumn{2}{c}{\textbf{Method}}
& \multicolumn{1}{c}{\makecell{\textbf{Aesthetic}\\\textbf{Quality}}}
& \multicolumn{1}{c}{\makecell{\textbf{Background}\\\textbf{Consistency}}}
& \multicolumn{1}{c}{\makecell{\textbf{Dynamic}\\\textbf{Degree}}}
& \multicolumn{1}{c}{\makecell{\textbf{Imaging}\\\textbf{Quality}}}
& \multicolumn{1}{c}{\makecell{\textbf{Motion}\\\textbf{Smoothness}}}
& \multicolumn{1}{c}{\makecell{\textbf{Subject}\\\textbf{Consistency}}} \\
\midrule

\multirow{4}{*}{General Model}
& CogVideoX
& \num{42.34}
& \num{92.59}
& \num{68.95}
& \num{62.68}
& \num{99.17}
& \num{85.27} \\

& Hunyuan
& \num{42.14}
& \bfseries \num{94.67}
& \num{8.42}
& \num{62.51}
& \bfseries \num{99.68}
& \bfseries \num{93.53} \\

& Wan~2.2
& \num{41.13}
& \num{90.63}
& \bfseries \num{92.11}
& \num{59.80}
& \num{98.15}
& \num{84.14} \\

& {\color[HTML]{363636}Ours-All}
& \bfseries \num{41.83}
& \num{91.95}
& \num{47.37}
& \bfseries \num{66.14}
& \bfseries \num{99.55}
& \num{89.06} \\

\midrule

\multirow{2}{*}{Physical Model}
& Force Prompting
& \num{38.38}
& \bfseries \num{97.54}
& \num{0.00}
& \num{61.90}
& \bfseries \num{99.77}
& \bfseries \num{96.15} \\

& Ours-Force
& \num{38.36}
& \num{91.94}
& \num{44.57}
& \num{65.00}
& \bfseries \num{99.57}
& \num{90.06} \\

\bottomrule
\end{tabular}%
}
\end{table*}

\noindent\textbf{Spatial Structural Stability (Multi-Granularity).}

(a)\noindent\textit{ Pixel-level MSE:}
Given a paired reference, the frame-wise mean squared error is
\begin{equation}
\label{eq:mse}
\mathrm{MSE}
=\frac{1}{T}\sum_{t=1}^{T}\frac{1}{|\Omega|}\sum_{p\in\Omega}\big(I_t(p)-\tilde I_t(p)\big)^2,
\end{equation}
where $I_t(p)$ and $\tilde I_t(p)$ are pixel intensities at location $p$.
Subject-/background-masked variants simply restrict the inner sum to $\{p\mid S_t(p)=1\}$ or $\{p\mid B_t(p)=1\}$.

(b)\noindent\textit{ Object-level Subject Consistency}~\textup{(OC):}
To measure object identity/structure stability across time, we compute the cosine similarity of masked features between adjacent frames:
\begin{equation}
\label{eq:sc}
\mathrm{OC}
=\frac{1}{T-1}\sum_{t=1}^{T-1}
\frac{\left\langle \phi(I_t\odot S_t),\, \phi(I_{t+1}\odot S_{t+1}) \right\rangle}
{\big\|\phi(I_t\odot S_t)\big\|_2\;\big\|\phi(I_{t+1}\odot S_{t+1})\big\|_2},
\end{equation}
where $\odot$ denotes element-wise masking.
Higher OC indicates more stable object appearance/structure.

(c)\noindent\textit{ Region-level IoU:}
For region geometry/trajectory agreement with a reference, we average the intersection-over-union across frames:
\begin{equation}
\label{eq:iou}
\mathrm{IoU}
=\frac{1}{T}\sum_{t=1}^{T}
\frac{|M_t\cap \hat M_t|}{|M_t\cup \hat M_t|},
\end{equation}
where $|\cdot|$ is set cardinality (pixel area). Higher is better.

(d)\noindent\textit{ Scene-level Background Consistency}~\textup{(SC):}
We assess frame-to-frame background stability (lighting/texture/geometry) with masked SSIM:
\begin{equation}
\label{eq:bc}
\mathrm{SC}
=\frac{1}{T-1}\sum_{t=1}^{T-1}
\mathrm{SSIM}\!\big(I_t\odot B_t,\; I_{t+1}\odot B_{t+1}\big),
\end{equation}
where $\mathrm{SSIM}(\cdot,\cdot)$ is the structural similarity index applied on background-only content.
(PSNR/LPIPS can be reported analogously; we use SSIM for bounded interpretability.)

\noindent\textbf{Semantic--Physical Alignment.}

\begin{figure*}[!htb]
    \centering
    \includegraphics[width=0.95\linewidth]{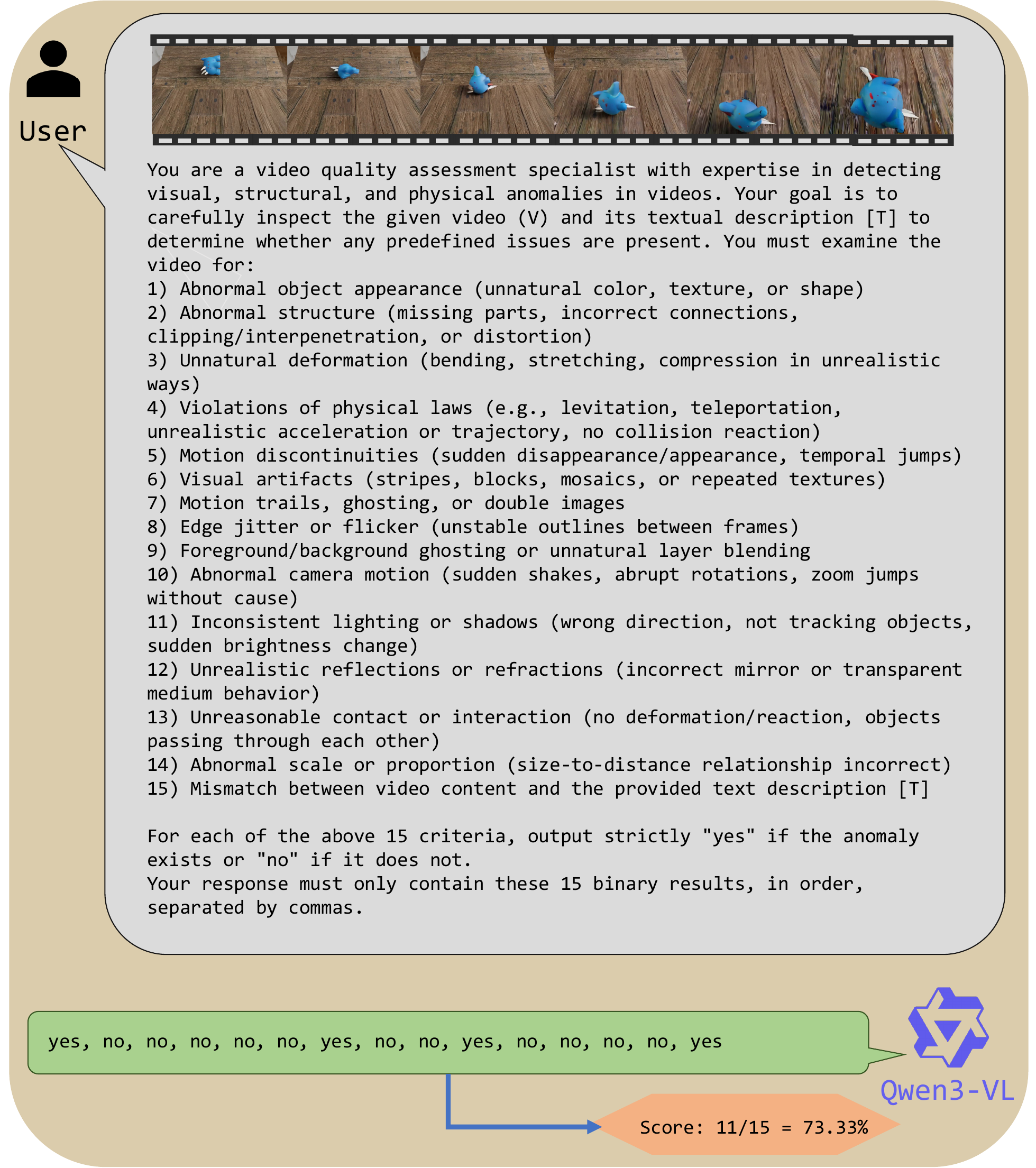} 
    \caption{
        An overview of the VLM-QA pipeline.
    }
    \label{fig:vlmqa_pipeline}
\end{figure*}

\noindent\textit{VLM-QA:} We evaluate semantic and physical plausibility with a physics-aware question set $\mathcal{Q}=\{q_k\}_{k=1}^{K}$ (e.g., penetration, gravity consistency, motion continuity, abnormal deformation).
A vision--language model (VLM) answers each $q_k$ on video $V$ (and its text prompt) and returns $a_k$.
Each answer is mapped to a binary indicator $g(a_k)\in\{0,1\}$, where $1$ means ``passes the criterion''.
The final score is
\begin{equation}
\label{eq:vlm}
\mathrm{VLM\mbox{-}QA}
=\frac{1}{K}\sum_{k=1}^{K} g(a_k).
\end{equation}
We also include a caption-matching question $[\mathrm{V}]\,\text{``Does this video match the description: ''}[\mathrm{T}]$. The overall scoring pipeline is illustrated in Figure~\ref{fig:vlmqa_pipeline}.

For each metric, we report the mean over videos. For $\mathrm{OC}$ and $\mathrm{SC}$ we additionally report the temporal standard deviation within each video to reflect intra-video stability.
When a paired reference is unavailable, we omit MSE/IoU and keep FVMD, OC/SC (intra-video), and VLM-QA.

\section{Human Evaluation Implementation Details}
\label{sec:appendix_human}

\begin{figure*}[!htb]
    \centering
    \includegraphics[width=0.95\textwidth]{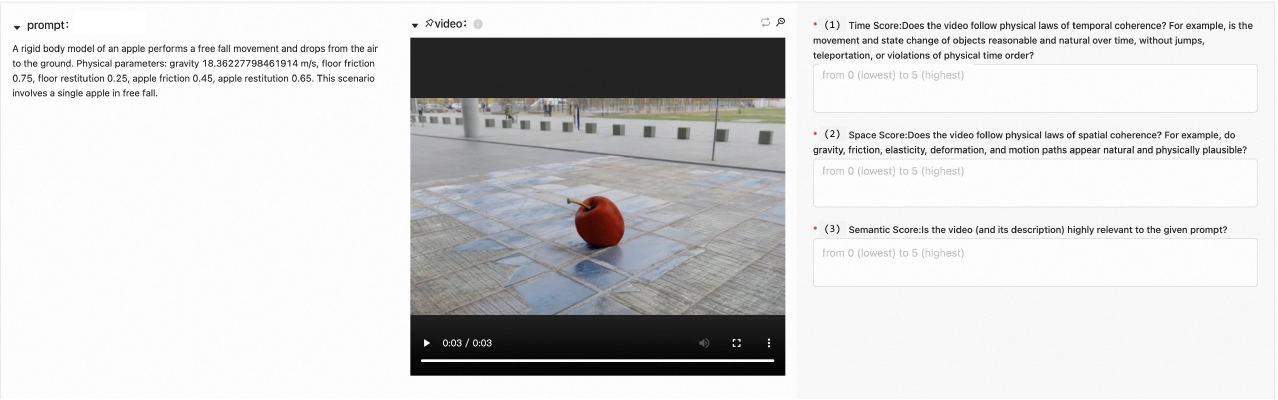}
    \caption{The interface used for user annotations in our study.}
    \label{fig:annotation_interface}
\end{figure*}

We divided all participants into five groups and uploaded the videos to an evaluation platform. Users were asked to rate each video on a scale from 0 to 5 across three dimensions: temporal consistency, spatial accuracy, and semantic coherence. The evaluation questions were formulated as follows:
\begin{tcolorbox}[colback=black!5!white, title=Video Evaluation — User Study Scoring Template, fonttitle=\bfseries, colframe=black!50!white, boxrule=0.8pt]
\small
Please watch the video and refer to the given description, then score according to the following three dimensions.  
Each dimension ranges from \textbf{0} (lowest) to \textbf{5} (highest), based on your subjective evaluation.

\begin{enumerate}[label=\arabic*)]
    \item \textbf{Time Score}  
    Does the video follow physical laws of temporal coherence? For example, is the movement and state change of objects reasonable and natural over time, without jumps, teleportation, or violations of physical time order?  
    \begin{itemize}
        \item \textbf{5 points}: Fully consistent with physical temporal coherence
        \item \textbf{0 points}: Not consistent, with obvious violations of temporal order or abrupt changes
    \end{itemize}

    \item \textbf{Space Score}  
    Does the video follow physical laws of spatial coherence? For example, do gravity, friction, elasticity, deformation, and motion paths appear natural and physically plausible?  
    \begin{itemize}
        \item \textbf{5 points}: Fully consistent with physical spatial laws; no violations like gravity defiance or implausible deformations
        \item \textbf{0 points}: Clearly violates spatial physical laws
    \end{itemize}

    \item \textbf{Semantic Score}  
    Is the video (and its description) highly relevant to the given prompt?  
    \begin{itemize}
        \item \textbf{5 points}: Highly consistent with the prompt; all important elements, actions, and relationships match
        \item \textbf{0 points}: Completely unrelated or severely diverging from the prompt
    \end{itemize}
\end{enumerate}

\vspace{5mm}
\noindent\textbf{Please fill in:}
\begin{verbatim}
Time Score (0-5): ____
Space Score (0-5): ____
Semantic Score (0-5): ____
\end{verbatim}

\end{tcolorbox}
\begin{figure}[htbp]
    \centering
    \includegraphics[width=\linewidth]{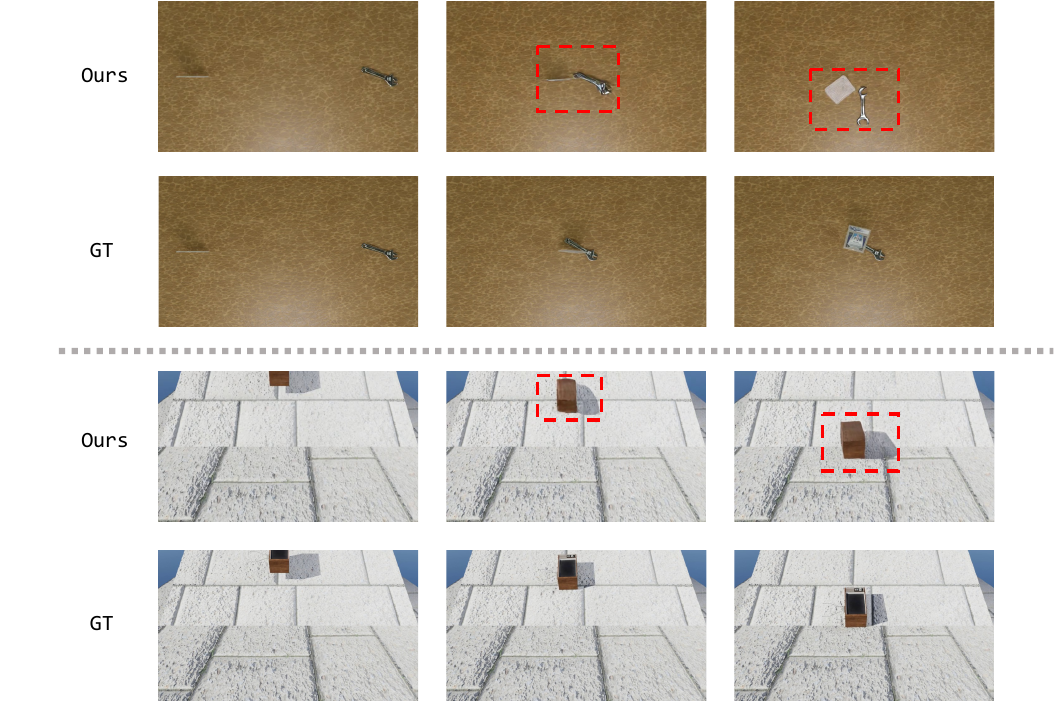}
    \caption{
        Failure cases.
    }
    \label{fig:failure_cases}
\end{figure}
The interface used for user annotations is shown in Figure~\ref{fig:annotation_interface}.

\section{More Experiments}
\label{sec:appendix_exper}

\noindent\textbf{Generalization on VBench Metrics.} We further evaluate our models on VBench to assess overall visual quality beyond physical consistency (Table~\ref{tab:vbench_metrics}).
Among general-purpose baselines, \textsc{Ours-All} attains the best Imaging Quality (66.14) and achieves competitive Motion Smoothness (99.55), suggesting that incorporating physics-aware learning does not compromise visual fidelity.
For physics-oriented settings, \textsc{Ours-Force} maintains strong perceptual quality (Imaging Quality = 65.00) and stable motion, while substantially outperforming Force Prompting in dynamic realism (Dynamic Degree = 44.57 vs.\ 0.00).
Overall, these results indicate that our method preserves aesthetic quality and temporal coherence while improving physical plausibility, demonstrating better generalization to real-world scenarios.

\noindent\textbf{Failure Cases.} As shown in the Figure\ref{fig:failure_cases}, our method fails in certain scenarios, primarily due to incomplete capture of the target object or its excessively small proportion within the overall scene. These conditions limit the acquisition of crucial shape features, making it difficult for the model to reliably infer the subsequent motion trajectories and predict the object's new states and appearances. Moreover, such scenarios occur infrequently in the training data, resulting in insufficient generalization ability for these cases. To address this limitation, future work could incorporate targeted data augmentation to increase the prevalence of such samples, or develop specialized modeling strategies for small-scale and partially visible objects, thereby enhancing the model's robustness and prediction accuracy in complex environments.